\documentclass{article}

\usepackage[utf8]{inputenc}
\usepackage{amsmath}
\usepackage{amssymb}
\usepackage{graphicx}
\usepackage{subcaption}
\usepackage{dsfont}  
\usepackage{authblk}
\usepackage{xr-hyper}
\usepackage{hyperref}
\usepackage{upgreek}
\usepackage{siunitx}
\DeclareSIUnit{\wtpercent}{wt\%}
\usepackage{bbm}
\usepackage{mathrsfs}
\usepackage[title]{appendix}
\usepackage{pdfpages}
\usepackage{algorithm}
\usepackage{algpseudocode}

\usepackage{float}
\usepackage{booktabs}
\usepackage{adjustbox}
\usepackage{tikz}
\usetikzlibrary{shapes, arrows, positioning,decorations.markings}
\tikzstyle{vecArrow} = [thick, decoration={markings,mark=at position
	1 with {\arrow[semithick]{open triangle 60}}},
double distance=1.4pt, shorten >= 5.5pt,
preaction = {decorate},
postaction = {draw,line width=1.4pt, white,shorten >= 4.5pt}]

\tikzstyle{redvecArrow} = [color=red,thick, decoration={markings,mark=at position
	1 with {\arrow[color=red,semithick]{open triangle 60}}},
double distance=1.4pt, shorten >= 5.5pt,
preaction = {decorate},
postaction = {draw,line width=1.4pt, white,shorten >= 4.5pt}]

\tikzstyle{innerWhite} = [semithick, white,line width=1.4pt, shorten >= 4.5pt]
\hypersetup{hidelinks}

\usepackage[tmargin=1in,bmargin=1in,lmargin=0.75in,rmargin=0.75in]{geometry}
\usepackage{upgreek}

\usepackage{xcolor}

\usepackage{manyfoot}
\SetFootnoteHook{\hspace*{-1.8em}}
\DeclareNewFootnote{bl}[gobble]
\usepackage[nolist,nohyperlinks]{acronym}

\usepackage{multirow}
\usepackage{makecell}

\usepackage[biblabel,nosort]{cite}

\def\submission{0}
\if\submission1
\usepackage[doublespacing]{setspace}
\usepackage{lineno}
\linenumbers
\fi

\makeatletter
\newcommand*{\addFileDependency}[1]{
	\typeout{(#1)}
	\@addtofilelist{#1}
	\IfFileExists{#1}{}{\typeout{No file #1.}}
}
\makeatother

\newcommand{\numSlices}{{n_{\mathrm{data}}}}

\newcommand{\R}{\mathbb{R}}
\newcommand{\Z}{\mathbb{Z}}

\newcommand{\field}{X}

\newcommand{\noise}{N}
\newcommand{\kernel}{k}
\newcommand{\window}{D}
\newcommand{\samplingWindow}{W}

\newcommand{\samplingWindowTwo}{\samplingWindow_{\mathrm{2D}}}
\newcommand{\samplingWindowThree}{\samplingWindow_{\mathrm{3D}}}

\newcommand{\expectation}[1]{\mathbb{E}\!\left[#1\right]}
\newcommand{\probability}[1]{\mathbb{P}\!\left(#1\right)}

\newcommand{\expectationFunction}{m}
\newcommand{\covarianceFunction}{\rho}

\newcommand{\fieldOne}{X}
\newcommand{\fieldTwo}{Y}

\newlength\imageLength
\newcommand{\fieldOneCorrelated}{X_\mathrm{c}}
\newcommand{\fieldTwoCorrelated}{Y_\mathrm{c}}

\newcommand{\lx}{\lambda_\mathrm{X}}
\newcommand{\ly}{\lambda_\mathrm{Y}}

\newcommand{\chiOne}{X^\prime}
\newcommand{\chiTwo}{Y^\prime}

\newcommand{\cathodeModel}{\Xi}

\newcommand{\cathodeModelCutout}{{\Xi}^{\mathrm{cutout}}}
\newcommand{\cathodeModelCutoutRealization}{{\xi}^{\mathrm{cutout}}}

\newcommand{\discriminator}{D}
\newcommand{\discriminatorWeights}{{\theta_\mathrm{D}}}
\newcommand{\discriminatorWeightsSet}{\Theta_\mathrm{D}}

\newcommand{\cathodeModelParametric}{\Xi_\theta}
\newcommand{\fittedParam}{\widehat{\theta}}
\newcommand{\fittedCathodeModel}{\cathodeModel_{\fittedParam}}

\newcommand{\anisotropicCathodeModel}{\cathodeModel^{\mathrm{ALP}}}

\newcommand{\NonParametricCathodeModel}{\cathodeModel^{\mathrm{HP}}}

\newcommand{\LowParametricCathodeModel}{\cathodeModel^{\mathrm{LP}}}

\newcommand{\cathodeModelParametricApprox}{\widetilde{\cathodeModel}_\theta}

\newcommand{\heaviside}{H}

\newcommand{\lr}{\mathrm{lr}}
\newcommand{\bs}{\mathrm{bs}}
\newcommand{\loss}{\mathrm{loss}}

\DeclareMathOperator*{\argmin}{arg\,min}

\newcommand{\ie}{\emph{i.e.}}
\newcommand{\eg}{\emph{e.g.}}

\title{Generative adversarial framework to calibrate excursion set models for the 3D morphology of all-solid-state battery cathodes}

\def\submissionmain{0}
\if\submissionmain1
\author{}
\else
\author{Orkun Furat$^{1,\dag,\ast}$,  Sabrina Weber$^{1,\dag}$, Johannes~Schubert$^{2}$, René~Rekers$^{2}$,  Maximilian~Luczak$^3$,  Erik~Glatt$^3$, Andreas~Wiegmann$^3$, Jürgen~Janek$^2$, Anja~Bielefeld$^2$, Volker~Schmidt$^1$}
\fi

\date{}

\begin{document}
\begin{acronym}
	\acro{SEM}{scanning electron microscopy}
	\acro{GAN}[GAN]{generative adversarial network}
	\acro{MSSIM}{mean structural similarity}
	\acrodefplural{GAN}[GANs]{generative adversarial networks}
	\acro{EBSD}{electron backscatter diffraction}
	\acro{NMC}{LiNi$_{1-x-y}$Mn$_y$Co$_x$O$_2$}
	\acro{CNN}{convolutional neural network}
\end{acronym}

\maketitle

\begin{center}
	\it
	$^1$Institute of Stochastics, Ulm University, Helmholtzstraße 18, 89069 Ulm, Germany
	\\
	$^2$Center for Materials Research (ZfM), Justus Liebig University Giessen, Heinrich-Buff-Ring 16, Giessen 35392, Germany
	\\
	$^3$Math2Market GmbH, Richard-Wagner-Straße 1, 67655 Kaiserslautern, Germany
\end{center}
\footnotebl{
\noindent
$^\ast$Corresponding author; \textit{Email address:} orkun.furat@uni-ulm.de (Orkun Furat)\\
$^\dag$ OF and SW contributed equally to this paper.
}

\begin{abstract}
	\noindent
This paper presents a computational method for generating virtual 3D morphologies of functional materials using low-parametric stochastic geometry models, \ie, digital twins, calibrated with 2D microscopy images. These digital twins allow systematic parameter variations to simulate various morphologies, that can be deployed for virtual materials testing by means of spatially resolved numerical simulations of macroscopic properties. Generative adversarial networks (GANs) have gained popularity for calibrating models to generate realistic 3D morphologies. However, GANs often comprise of numerous uninterpretable parameters make systematic variation of morphologies for virtual materials testing challenging. In contrast, low-parametric stochastic geometry models (\eg, based on Gaussian random fields) enable targeted variation but may struggle to mimic complex morphologies. Combining GANs with advanced stochastic geometry models (\eg, excursion sets of more general random fields) addresses these limitations, allowing model calibration solely from 2D image data. This approach is demonstrated by generating a digital twin of all-solid-state battery (ASSB) cathodes. Since the digital twins are parametric, they support systematic exploration of structural scenarios and their macroscopic properties. The proposed method facilitates simulation studies for optimizing 3D morphologies, benefiting not only ASSB cathodes but also other materials with similar structures.

\end{abstract}

\textbf{\emph{Keywords:}}---Spatial stochastic model, stereology,  generative adversarial network, all-solid-state battery cathode, microscopic image data.


\section{Introduction}\label{sec:introduction}

\noindent
It is consensus that the properties of (functional) materials for batteries, such as the accessible charge capacity or ionic conductivity, are significantly influenced by their  nano/microstructure 
\cite{muller2017investigation,ASHERI2024110370,vu2023towards,McLaughlin20223,Marmet_2024}. These structure-property relationships are of crucial importance to provide structuring recommendations for materials with optimized performance \cite{stenzel17,Prifling_2021}. However, such recommendations, by themselves, are often not sufficient---one must also know how to manufacture materials with  desired structures. For this purpose, experimental variations of manufacturing processes can be used to produce materials with different structures and compositions. Subsequently, the influence of parameters of the manufacturing processes on the   nano/microstructure of materials, \ie, so-called process-structure relationships can be investigated \cite{alabdali2023three}. In order to derive both types of relationships it is necessary to quantitatively characterize the nano/microstructures of materials,  using, \eg, structural descriptors (like porosity, specific surface area, etc.) \cite{ohser2000statistical}. Imaging techniques, such as computed electron or X-ray tomography (nano-CT or micro-CT), or focused ion beam (FIB) milling followed by scanning electron microscopy (SEM), allow for a quantitative characterization of the 3D nano/microstructure of materials, followed by   statistical analysis and modeling of image data \cite{withers2007,maire14,BURNETT2016119}. In particular, regression models can be deployed to predict the nano/microstructure of materials for any feasible configuration of process parameters \cite{prifling2019parametric}. 
This enables the calibration of process parameters by means of computer-aided optimization \cite{furat20} such that, according to resulting structuring recommendations, structures with improved  properties can be manufactured. In order to derive such  process-structure-property relationships an important step is to derive a quantitative mathematical description of materials nano/microstructure. 

Particularly, stochastic geometry \cite{Chiu2013} provides useful mathematical tools for describing the nano/microstructures of various materials by means of spatial stochastic modeling  
\cite{Furat2021,NEUMANN2023112394,kuchler2018,JUNG2024110474,jeziorski2024stochasticgeometrymodelstexture}. 
More precisely, once calibrated to (\eg, 3D image) data, these kinds of stochastic models can be regarded as digital twins for material nano/microstructures since they allow for the generation of virtual, but realistic 3D nano/microstructures. Consequently, stochastic geometry models are informative mathematical tools for characterizing the nano/microstructure of materials with various applications.
For example,  using 3D morphologies generated by stochastic geometry models as input for spatially resolved numerical simulations (\eg, tensile tests or transport processes), comprehensive  databases of morphologies and effective macroscopic material properties can be determined by means of computer simulations \cite{Prifling_2021,Foehst2022,Daubner2024}.
For a wide range of applications, such as batteries and fuel cells, virtual material microstructures can be generated stochastically and their properties can be analyzed using the specialized software tool GeoDict, which enables the simulation and evaluation of electrode properties, such as charging behavior, to support material optimization and performance assessment \cite{SANDHERR2023107359,Wenzler_2023,Clausnitzer2023}. 
It should be emphasized here that parametric stochastic geometry models, in particular because of their interpretable model parameters, are extremely valuable. 
Before deploying a parametric model, typically it has to be investigated whether an adequate model type has been chosen for  describing experimentally measured data. Therefore, in the case of stochastic geometry models a common approach is to optimize model parameters until the discrepancy between generated structures and data is as small as possible \cite{THEODON2024119983}. If this optimization problem delivers satisfactory results, it is assumed (possibly after some further validation steps) that the parametric model can be used to describe the data. Otherwise, the type of the parametric model would have to be modified, \eg, by increasing its complexity. 

Once it is ensured that realistic  nano/microstructures can be generated by 
a parametric stochastic geometry model, it can be deployed for investigating further structural scenarios that have not been yet observed within data. More precisely, a systematic variation of model parameters allows for the generation of a wide range of structurally different morphologies followed by the simulation of their corresponding macroscopic properties. This means that parametric models enable the generation of large databases comprised of nano/microstructures together with their macroscopic properties \cite{Prifling_2021,neumann2020quantifying}.
Consequently, this type of virtual materials generation and testing can be utilized to derive structure-property relationships while reducing the number of real-life experiments 
(\eg, material synthesis, imaging, analysis of macroscopic properties) necessary to derive similar relationships \cite{stenzel17,Prifling_2021,neumann2020quantifying,MOUSSAOUI2018262,Barman2019}.

Stochastic geometry offers models that capture a wide range of morphologies, \eg, line patterns, random graphs, packings of  particles and complex tessellations of space \cite{Chiu2013,jeulin2021morphological}.
For multiphase materials, \eg, for porous media a commonly considered model type are excursion sets of so-called random fields \cite{Chiu2013,adler2009random}. More precisely, random fields are stochastic models that can be used to randomly generate  functions that assign points of Euclidean space with a random number. Then, the morphology of material phases can be modeled by considering all points in space for which the random value is above a certain threshold---resulting in the excursion set.
Well-studied cases of this class of models are excursion sets of Gaussian random fields or $\chi^2$ random fields \cite{Chiu2013,Neumann2024Morphology}. For these models there are some analytical formulas that allow for the calibration of model parameters to 3D image data, \ie, these formulas ensure that parameters of excursion set models are chosen such that generated morphologies exhibit similar statistics as nano/microstructures observed in 3D image data.
Thus, choosing model parameters that minimize the discrepancy between generated structures and imaged ones is quite forward, however, for these ``simple parametric models'' it is not always ensured that such a minimum enables the generation of realistic morphologies. In other words, the model type may not be sufficient for representing the data.
However, as model complexity increases---such as for excursion sets of more general random fields, which are necessary to capture more intricate nano/microstructures---the required number of model parameters increases substantially. This can make classical model calibration by means of interpretable descriptors impractical. 


In recent years, non-parametric methods of machine learning have emerged that can be calibrated to data without possibly restrictive model assumptions, allowing for the generation of statistically similar samples as observed in the data \cite{zheng2024text,nguyen2022synthesizing,gayon2020pores}. Typically, these data-driven models are referred to as generative artificial intelligence (AI). Rather established examples from generative AI are generative adversarial networks (GANs) \cite{Goodfellow2014}. In a classical setup, GANs consist of two competing networks, referred to as generator and discriminator. During training, the task of the generator (which receives noise as input) is to generate random samples that are statistically similar to the training data. In other words, the generator is supposed to model the (often multivariate) probability distribution associated with the data. The discriminator is trained to distinguish between  ``real data'' and samples generated by the generator. For training purposes, the generator receives feedback of the discriminator's decision such that its weights are iteratively improved, in order to ``trick'' the discriminator. GANs or similar generative networks, which have been successfully trained with 3D image data of a materials' nano/microstructure, can be considered to be calibrated stochastic geometry models / digital twins, since they allow for the generation of virtual but realistic morphologies. One significant advantage of these non-parametric approaches is that they need almost no assumptions on the morphologies to be modeled, \ie, the same network architectures can be used to model a broad range of morphologies. In addition, non-parametric methods have been used to calibrate 3D models based on 2D image data, which is often a non-trivial problem \cite{kench2021generating,phan2024generating,BHADURI2021110709}. For example, in \cite{kench2021generating}  GANs have been trained to generate random 3D images, the planar sections of which  are statistically similar to 2D image data. 
However, methods from generative AI can have some drawbacks in comparison to conventional parametric approaches of stochastic geometry. For example, the generators deployed in GANs are neural networks, the parameters (\ie, the weights) of which are often not interpretable. Consequently, it is not straightforward to systematically vary generator parameters for investigating a broad range of structural scenarios to perform the virtual materials testing approach outlined above.

In the present paper, we introduce a computational method for using 2D image data to calibrate parametric stochastic geometry models for the 3D multiphase morphology of (functional) materials. More precisely, we consider rather flexible excursion sets of generalizations of Gaussian/$\chi^2$ random fields. For the parametric model considered in the present paper, there are no analytical formulas that allow for the direct calibration to image data.
To overcome this limitation, we combine our parametric stochastic geometry models with GANs for the purpose of model calibration \cite{fuchs2024generating}.
The considered models are flexible enough to stochastically model complex 3D morphologies, enabling the systematic exploration of different structures. Thus, they can also be deployed for virtual materials testing purposes which is not straightforward when deploying GANs only. 
The method described in the present paper will be used to parametrically model the three-phase microstructure of cathodes in all-solid-state batteries (ASSBs)---that might offer increased energy density and safety in comparison to commercially available batteries \cite{janek2023challenges}. Therefore, the presented digital twin, \ie, the parametric stochastic 3D model for the 3D morphology of cathodes in ASSBs can serve as a starting point for virtual materials testing \cite{neumann2020quantifying,KimDigitalTwin}.

\section{Methods}\label{sec:methods}

\subsection{Description of material and data acquisition}\label{sec:material}


\noindent
In this section we shortly describe the ASSB cathode material and the corresponding microscopy image data, on which the proposed method for the calibration of digital twins is demonstrated. A detailed description on cathode preparation, solid electrolyte (SE) synthesis, imaging and segmentation is provided in \cite{Minnmann_2024}. 
The ASSB cathodes considered in the present paper are comprised of glassy Li$_3$PS$_4-0.5$LiI and LiNi$_{0.83}$Mn$_{0.06}$Co$_{0.11}$O$_2$ which serve as solid electrolyte (SE) and active material (AM) within the cathode, respectively. In order to obtain SE particles with a sufficiently small particle size, initially synthesized SE particles have been comminuted in a planetary ball mill (Pulverisette 7,
Fritsch, Germany) with milling media of
\SI{3}{\milli\meter} diameters, see \cite{Minnmann_2024} for further details.
After fabricating the cathode with these SE and AM particles, the microstructure of the resulting ASSB cathode was imaged by SEM. To obtain a 3D tomogram, slices of the electrode were cut by plasma focused ion beam (PFIB) followed by SEM imaging. 
The resulting 3D image data have been segmented into three phases, \ie, the SE and AM phase, and a phase associated with remaining porosity within the cathode material, see \cite{Minnmann_2024} for details regarding image processing. After one-hot encoding the  experimentally measured 3D image data can be considered to be a mapping $S\colon \samplingWindowThree \times \{1,2,3\} \to \{0,1\}$, which is given by
\begin{equation}
	S((x,y,z),i) =
	\left\{
	\begin{array}{ll}
		1, & \text{if $i=1$ and $(x,y,z)$ is associated with the SE phase,}\\
		1, & \text{if $i=2$ and $(x,y,z)$ is associated with the AM phase,}\\
		1, & \text{if $i=3$ and $(x,y,z)$ is associated with the pore space,}\\
		0, & \text{else,}
	\end{array}
	\right.
	\label{eq:segmentation}
\end{equation}
for each voxel $(x,y,z)$ in some cuboidal sampling window $\samplingWindowThree = \{1,\dots, n_\mathrm{x}\} \times \{1,\dots, n_\mathrm{y}\} \times \{1,\dots, n_\mathrm{z}\}\subset \Z^3$ and for each channel $i=1,2,3$. The integers $n_\mathrm{x},n_\mathrm{y},n_\mathrm{z}>0$ denote the side lengths of the cuboidal sampling window.

Note that during fabrication of the cathode composite, a powder mixture that is comprised of SE and AM particles is pressed into a cylindrical cell setup which forms the cathode half-cell. Unlike in wet-processed electrodes, no conductive carbon or binders are added.
In particular, during cathode pressing uniaxial stress has been deployed. Consequently, microstructural statistics determined with respect to specific directions (\eg, directional shortest path lengths) may depend on the orientation of the chosen coordinate system. Therefore, we cannot assume that the microstructure of ASSB cathodes can be described by a stochastic 3D model that is statistically invariant under rotations of the coordinate system---a simplifying property known as isotropy.

\begin{figure}[ht]
	\centering
	\begin{subfigure}[b]{0.415\textwidth}
		\centering
		\includegraphics[width=\textwidth]{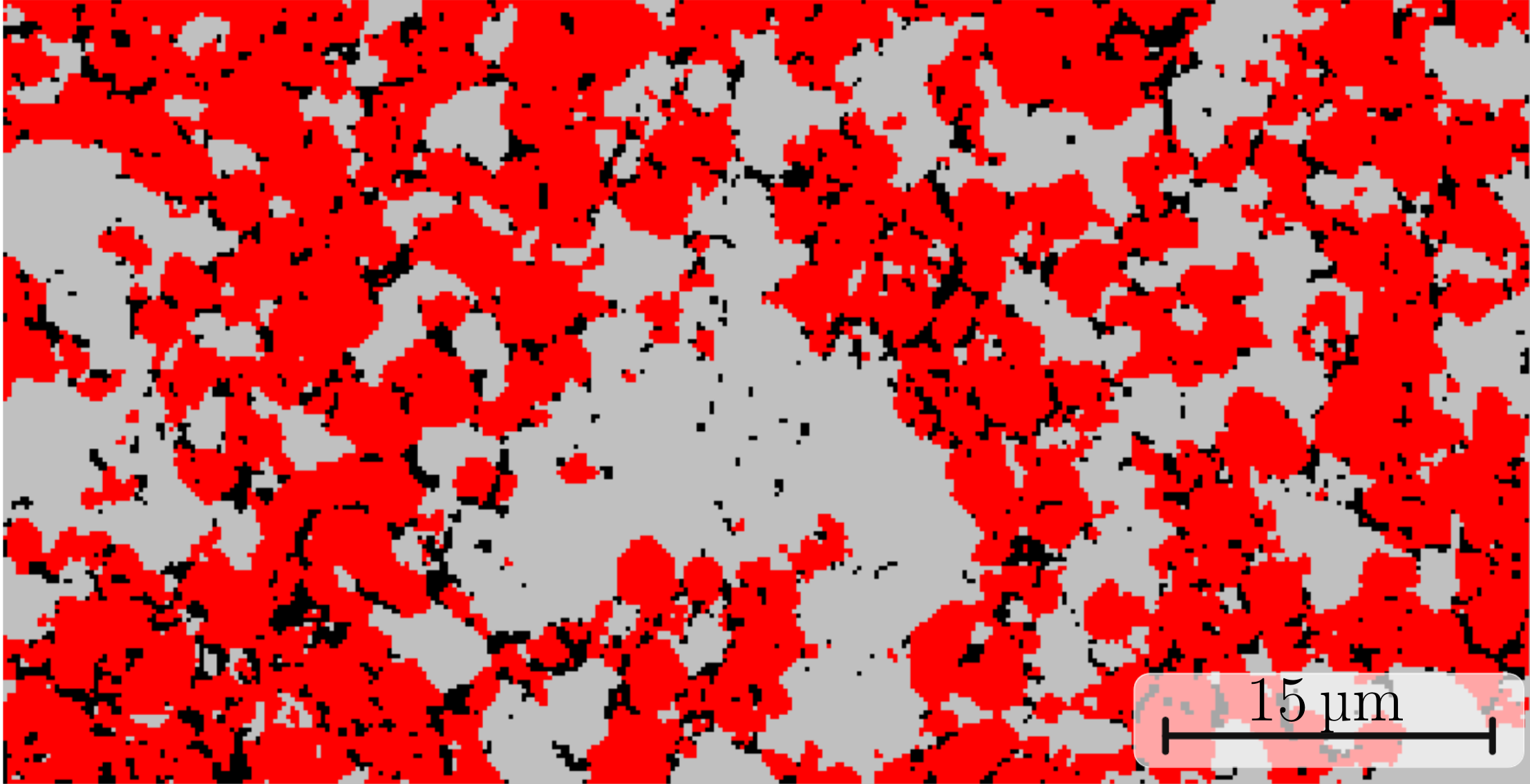}
		\caption{}		
	\end{subfigure}
	\hspace{1cm}
	\begin{subfigure}[b]{0.4\textwidth}
		\centering
		\includegraphics[width=\textwidth]{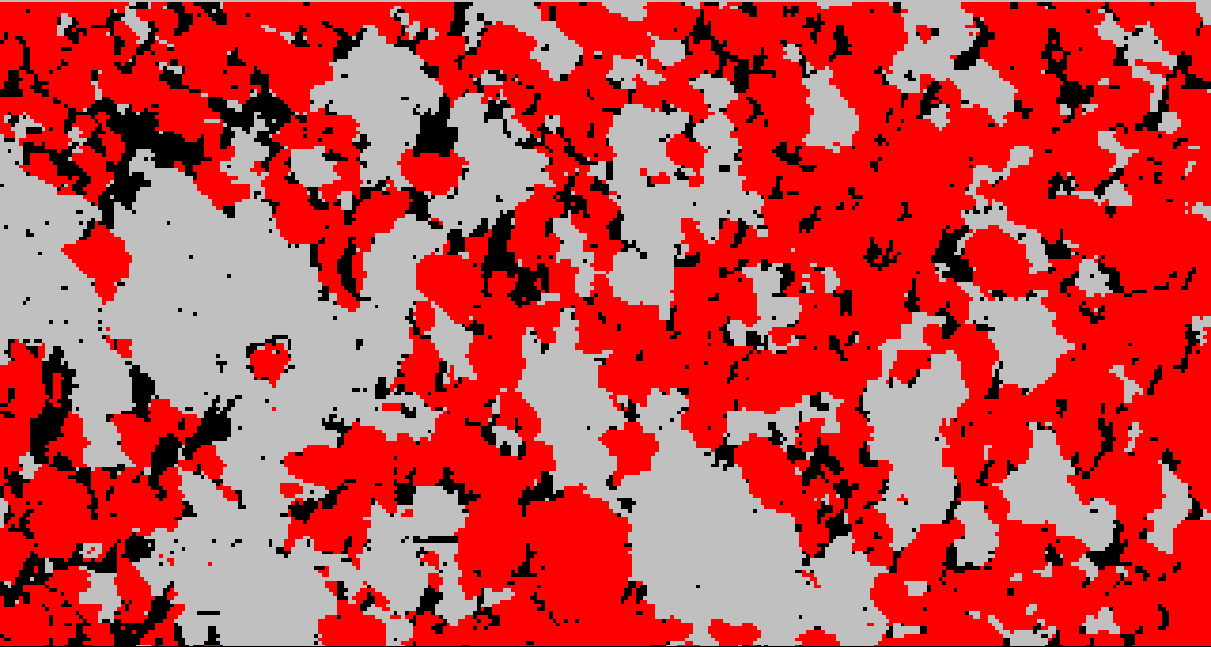}
		\caption{}		
	\end{subfigure}
	\caption{2D sections of segmented (experimentally measured) 3D image data, parallel to the $y$--$z$ plane  (a) and the $x$--$y$ plane (b). Both sub-figures use the same length scale. The pore space, active material and the solid electrolyte are represented by
		black, red and gray color, respectively.}
	\label{fig:anistropy}
\end{figure}

For the ASSB cathode specimen considered in this study, the uniaxial direction in which force has been applied during pressing aligns with the $z$-direction of the segmented image data represented by the mapping $S$ given in Eq.~\eqref{eq:segmentation}. Figure~\ref{fig:anistropy} visualizes the 2D segmentation along two perpendicular planar sections which are parallel 
to the $x$--$y$-plane and the $y$--$z$ plane. A first visual inspection suggests that both planar sections look statistically quite dissimilar which indicates anisotropy. Still the planar section parallel to the $x$--$y$ plane shown in Figure~\ref{fig:anistropy}b indicates isotropy with respect to rotations around the $z$-axis, \ie, cylindrical isotropy, which is also a reasonable assumption in the context of cathode production.

To develop an anisotropic, yet cylindrically isotropic spatial stochastic model, we  first derive an isotropic model in Section~\ref{sec:spatialstochastic}.
In particular, this means that arbitrarily oriented planar sections,   taken from 3D morphologies generated by the model, will look statistically similar to planar sections 
that are parallel to the $x$--$y$-plane.
Therefore, for the purpose of fitting the isotropic model, solely 2D image data is required, \ie, the model is fitted by a sequence of $\numSlices$ planar 2D images $S_1, \dots, S_\numSlices \colon \samplingWindowTwo \times \{1,2,3\} \to \{0,1\}$, where $\numSlices = n_\mathrm{z}$, which are given by
\begin{equation}\label{eq:slices}
	S_z((x,y),i) = S((x,y,z),i),
\end{equation}
for any $(x,y)\in \samplingWindowTwo= \{1,\dots, n_\mathrm{x}\} \times \{1,\dots, n_\mathrm{y}\}$, $z \in \{1,\dots, n_\mathrm{z}\}$ and $i\in\{1,2,3\}$.
Thus, in addition to modeling ASSB cathode microstructures, the methods described in Sections~\ref{sec:spatialstochastic} and \ref{sec:modelCalibration} below can also be applied in other situations, where only 2D image data of materials nano/microstructures is available, \eg, when acquiring 3D data is prohibitively expensive in terms of time and resources.

Finally, in a second step, in order to obtain a cylindrically isotropic model for the microstructure of ASSB cathodes, we locally stretch/compress the generated morphologies in $z$-direction, see Section \ref{sec:Cylindircallyisotropic}.

\subsection{Isotropic spatial stochastic modeling}\label{sec:spatialstochastic}
In this section we introduce some isotropic spatial stochastic models, the discretized realizations of which are mappings $\xi \colon \samplingWindow \times \{1,2,3\} \to \{0,1\}$ on some cuboidal sampling window $\samplingWindow\subset \Z^d$, where $d$ is some positive integer (\eg, $d=2$ or $d=3$), \ie, the mappings $\xi$ can be considered to be simulated image data of three-phase materials. Then, in Section~\ref{sec:modelCalibration}, we will describe methods for calibrating the models introduced in the present section to (2D) image data. The calibration aims to generate images that are statistically similar to the planar sections of the  segmented 2D image given by Eq.~(\ref{eq:slices}). 
After calibration of an isotropic model to 2D image data, a compression of the coordinate system along the $z$-axis will be performed in Section \ref{sec:Cylindircallyisotropic}, to obtain an anisotropic, yet cylindrically isotropic model.
For modeling purposes, we utilize so-called excursion sets of random fields. Therefore, in Section~\ref{sec:random_fields} we introduce random fields with a focus on Gaussian random fields which are well-studied models that can be used as building blocks for more complex models. Then, in Section~\ref{sec:excursionSets} we formally introduce the notion of excursion sets. Finally, in Section~\ref{sec:cathodeModel} a  flexible excursion set model is defined that can be used for modeling the 3D microstructure of ASSB cathodes.

\subsubsection{Random fields}\label{sec:random_fields}
As mentioned above, we utilize so-called random fields for modeling purposes. A common definition of random fields 
is that they are collections $\{\field(t)  \colon t \in T\}$ of random variables, where $T$ denotes some index set. A typical choice of $T$, when considering spatially continuous models, is the multi-dimensional Euclidean space, \ie, $T=\R^d$ for some integer $d>1$ denoting the dimension. However, in the present paper, due to computational reasons  we solely consider discretized versions of random fields. In particular, we choose $T=\Z^d$. 

\medskip\noindent
\textbf{Gaussian random fields.} A well-studied class of random fields is given by Gaussian random fields (GRFs), \ie, a random field $\{\field(t)  \colon t \in \Z^d\}$ of is called a GRF if the random vectors
$(\field(t_1),\dots, \field(t_n))$ are normal distributed for any $n \geq 1$ and $t_1\dots,t_n \in \Z^d$. Note that 
GRFs are uniquely characterized by their mean-value function $\expectationFunction \colon \Z^d \to \R$ and covariance function 
$\covarianceFunction \colon \Z^d \times \Z^d \to \R$, which are given by
\begin{equation}
	\label{mea.val.fun}\expectationFunction(t) =\expectation{X(t)}
\end{equation}
and
\begin{equation}
	\label{cov.var.fun}\covarianceFunction(s,t) = \mathrm{Cov}\!\left( X(s), X(t) \right)
	= \expectation{ \left(X(s) - m(s)\right)  \left(X(t) - m(t)\right) },
\end{equation}
for any $s,t \in \Z^d$.

\medskip\noindent
\textbf{Stationarity.} When  stochastically modeling the microstructure of  homogeneous materials (without structural gradients), a common model assumption is stationarity, where a random field is called stationary if its distribution is invariant with respect to translations. More precisely, a random field $\{\field(t)  \colon t \in \Z^d\}$ is called stationary if the distributional equality $(\field(t_1+ x),\dots, \field(t_n+x )) \stackrel{\mathrm{d}}{=} (\field(t_1),\dots, \field(t_n))$ holds for any  $n \geq 1$, $t_1\dots,t_n\in \Z^d$ and for each shift vector $x\in \Z^d$. 
In particular, for stationary random fields, the values  $\expectationFunction(t)$  of the mean-value function given in Eq.~\eqref{mea.val.fun}
are constant, and the values $\covarianceFunction(s,t)$ of the covariance function given in Eq.~\eqref{cov.var.fun}
only depend on the difference $t-s$, \ie, it holds that $\covarianceFunction(s,t)= \covarianceFunction(o, t-s)$ for any $s,t\in\Z^d$, where $o \in \Z^d$ denotes the origin. Thus, by abuse of notation, covariance functions of stationary random fields can be considered to be functions $\rho \colon \Z^d \to \R$ that are given by
$    \rho(t) = \rho(o,t)$. 
Note that in the following, we solely consider stationary random fields that are centered, \ie, for which $m(t)= 0$ holds for each $t\in\Z^d$.

A simple way to simulate (centered stationary) GRFs is given by considering moving averages of a special class of GRFs, so-called Gaussian white noise denoted by $\{\noise(t),t\in\Z^d\}$, where the random variable $\noise(t)$ is standard normal distributed (shortly: $\noise(t) \sim \mathcal{N}(0,1)$) for each $t\in \Z^d$, and the random variables $\noise(t_1)$ and $\noise(t_2)$ are independent if $t_1 \neq t_2$. Now, let $\kernel \colon \Z^d \to \R$ be a function such that $k(t)=0$ for each $t\not\in D$ and some bounded set $\window \subset \Z^d$,   referred to as kernel in the following.
Then, the moving average $\{\field(t)  \colon t \in \Z^d\}$  given by
\begin{equation}\label{eq:moving}
	\field(t) = \sum_{s\in \Z^d} \kernel(s) \, \noise(t-s) = (\noise \ast \kernel) (t),
\end{equation}
for each $t\in\Z^d$, is a centered stationary GRF, whose covariance function $\rho$ is given by
\begin{equation}\label{eq:kernel2cov}
	\rho(s,t) = (\kernel \ast \kernel_\mathrm{mirrored})(t-s),
\end{equation}
for any $s,t\in\Z^d$, where $\ast$ denotes convolution and $\kernel_\mathrm{mirrored}$ is the mirrored version of $\kernel$, \ie, $\kernel_\mathrm{mirrored}(t) = \kernel(-t)$ for each $t\in\Z^d$.
Note that the GRF given in Eq.~\eqref{eq:moving} 
is normalized, \ie, $\field(t) \sim \mathcal{N}(0,1)$ for each $t\in\Z^d$, if the kernel $\kernel$ or the covariance function $\covarianceFunction$ are normalized such that
\begin{equation}\label{eq:norm}
	\sum_{t \in D} \kernel^2(t)   =1
	\qquad\text{ or }
	\qquad\covarianceFunction(s,s)=1 \text{ for each } s \in \Z^d.
\end{equation}

On the other hand, it is possible to simulate stationary GRFs when the covariance function, but not the kernel, is known. More precisely, for a given covariance function $\rho\colon \Z^d \to \R$, we can determine the underlying kernel by
\begin{equation}\label{eq:Cov2Kernel}
	\kernel  = \mathrm{FFT}^{-1}\!\left( \sqrt{ \mathrm{FFT}(\rho) } \right),
\end{equation}
where we assume symmetry of the kernel $\kernel$, and $\mathrm{FFT}, \mathrm{FFT}^{-1}$ denote the (fast) Fourier transform and its inverse, respectively  \cite{abdallah2016morphological}.
Then, by deploying this kernel in Eq.~\eqref{eq:moving} we can simulate a GRF with covariance function $\rho.$ Note, however, that the inversion scheme given in Eq.~\eqref{eq:Cov2Kernel}  does not ensure that the computed kernel is real-valued nor that it has bounded support $D$. In practical applications, this issue can be addressed by considering the real part of the possibly complex-valued kernel $k$ and by constraining its support. Then, GRFs simulated with this real-valued kernel (with bounded support) exhibit a covariance function that approximates $\rho$.

\medskip\noindent
\textbf{Isotropy.} Besides stationarity,  another property of interest for modeling purposes is isotropy, \ie, distributional invariance with respect to rotations around the origin. 
The notion of isotropy can be easily defined on the continuous  Euclidean space $\R^d$, but not on the discretized index set $\Z^d$ as rotations of the grid do not necessarily result in the same grid.
Therefore, we first  mention some properties of isotropic random fields on Euclidean spaces, \ie, for $T=\R^d$, and discuss how they can be transferred to random fields on $\Z^d$ to ``approximate isotropy''.

If a random field on $\R^d$ is both  stationary and isotropic, its covariance function 
$\rho:\R^d\times\R^d\to\R$
does only depend on the distance between the considered points $s,t \in \R^d$, \ie,
we have 
\begin{equation}\label{eq:covIso}
	\rho(s,t) = \rho( |s-t| \, e_1),
\end{equation}
where $e_1=(1,0,\dots,0) \in \R^d$ is the unit vector parallel to the first axis and $|\cdot|$ denotes the Euclidean norm in $\R^d$. Therefore, in the case of stationary and isotropic random fields on $\R^d$, covariance functions are considered to be maps $\rho \colon [0,\infty) \to \R$. With an abuse of notation, the corresponding map $\rho \colon \R^d \times \R^d \to \R$ can be constructed by 
$\rho(s,t) = \rho(|s-t|)$,
for any $s,t \in \R^d.$ 
Similarly, in the case of  stationarity and isotropy,  for each $t\in\R^d$ the value $k(t)$ of the kernel $k:\R^d\to\R$ only
depends on the Euclidean norm $|t|$ of $t$. 
Therefore, for  discretized random fields with $T=\Z^d$ considered in the present paper, we can approximate this scenario by parameterizing kernels, \ie, by considering
\begin{equation}\label{eq:kernelNonNormalized}
	\kernel_{\alpha,\mathrm{non-normalized}} (t) =
	\left\{
	\begin{array}{ll}
		\alpha_{\left\lfloor |t| \right\rceil}, & \text{if }  {\left\lfloor |t| \right\rceil}\leq L,\\%
		0 & \text{else,}
	\end{array}
	\right.
\end{equation}%
for each $t\in \Z^d$, where ${\left\lfloor \cdot  \right\rceil}$ denotes the  nearest integer function and $\alpha=(\alpha_0, \dots,\alpha_L) \in \R^{L+1}$ for some $L\geq 0$. Consequently, the vector $\alpha$ can be considered to be a model parameter which characterizes stationary and ``approximately isotropic''  GRFs. For the sake of readability, we call a GRF isotropic if it is a moving average of Gaussian white noise with a kernel given by Eq.~(\ref{eq:kernelNonNormalized}).
Note that, by means of Eqs.~(\ref{eq:norm}) and (\ref{eq:kernelNonNormalized}) we can derive a parameterization $\kernel_\alpha:\Z^d\to\R$ of normalized kernels, where 
\begin{equation}\label{eq:isotropicKernel}
	\kernel_\alpha(t) = \frac{\kernel_{\alpha,\mathrm{non-normalized}}(t)}{
		\sqrt{ \sum_{s \in D}
			\kernel_{\alpha,\mathrm{non-normalized}}^2 (s)
		}
	},
\end{equation} 
for each $t\in \Z^d$. Then, the kernel $\kernel_\alpha$ characterizes a normalized stationary and isotropic GRF.
Unless stated otherwise, from here on we only consider normalized stationary and isotropic GRFs which are parameterized by means of Eq.~(\ref{eq:isotropicKernel}). These random field models will serve as building block for more complex models.

\medskip\noindent
\textbf{Correlated GRFs.} By means of Eq.~(\ref{eq:moving}) we can generate independent GRFs, by generating independent copies of Gaussian white noise $\noise$, followed by convolution with their respective kernels. However, in some applications it is desirable to generate correlated GRFs $\fieldOneCorrelated$ and $\fieldTwoCorrelated$, \eg, for modeling attraction or repulsion behavior of phases. The simulation of correlated GRFs can be easily achieved by combining independent GRFs.
For example, besides considering independent GRFs $X,Y$, we can additionally introduce three further independent GRFs $\widetilde{X}, \widetilde{Y}, \widetilde{Z}$
with covariance functions $\rho_{\widetilde{X}},\rho_{\widetilde{Y}},\rho_{\widetilde{Z}}$ to construct correlated GRFs.
More precisely, the random fields $\fieldOneCorrelated$ and $\fieldTwoCorrelated$ given by
\begin{equation}\label{eq:correlatedFields}
	\fieldOneCorrelated= \sqrt{1-\gamma} \, \widetilde{X} + \sqrt{\gamma} \, \widetilde{Z}
	\quad \text{and} \quad 	\fieldTwoCorrelated= \sqrt{1-\gamma}   \, \widetilde{Y} \pm \sqrt{\gamma}  \, \widetilde{Z}
\end{equation}
are normalized GRFs, which in general are not independent\footnote{Depending on the choice of the sign $\pm$ in Eq.~(\ref{eq:correlatedFields}), the random fields $\fieldOneCorrelated$ and $\fieldTwoCorrelated$ are either positively or negatively correlated.}, where $\gamma \in [0,1]$ is a model parameter. In particular, the  covariance functions $\rho_{\fieldOneCorrelated}$ and $\rho_{\fieldTwoCorrelated}$ of $\fieldOneCorrelated$ and $\fieldTwoCorrelated$ are given by
\begin{equation}
	\rho_{\fieldOneCorrelated} (s,t) = (1-\gamma)  \, \rho_{\widetilde{X}}(s,t) +\gamma \, \rho_{\widetilde{Z}}(s,t) \quad
	\text{and} \quad
	\rho_{\fieldTwoCorrelated} (s,t) = (1-\gamma) \, \rho_{\widetilde{Y}}(s,t) + \gamma \, \rho_{\widetilde{Z}}(s,t),
\end{equation}
for any $s,t \in \Z^d$. 
Since the variance of $\fieldOneCorrelated(s)$ and $\fieldTwoCorrelated(s)$ is given by $\rho_{\fieldOneCorrelated} (s,s)= \rho_{\fieldTwoCorrelated} (s,s)=1$, they are normalized. Furthermore, they are non-independent, as their joint covariance function $\rho_{\fieldOneCorrelated\fieldTwoCorrelated} \colon \Z^d \times \Z^d \to \R$ is given by
\begin{equation}
	\rho_{\fieldOneCorrelated\fieldTwoCorrelated}(s,t)
	=\expectation{
		\fieldOneCorrelated(s)\, \fieldTwoCorrelated(t) 
	}	
	=\pm  \gamma \rho_{\widetilde{Z}}(s,t),
\end{equation}
for any $s,t \in \Z^d$. 
In particular, we have 
$\rho_{\fieldOneCorrelated\fieldTwoCorrelated}(t,t) = \expectation{\fieldOneCorrelated(t) \, \fieldTwoCorrelated(t)} = \pm \gamma$,
for each $t\in \Z^d$, \ie, $\gamma$ is the absolute value of the covariance between the random variables $\fieldOneCorrelated(t)$ and $\fieldTwoCorrelated(t)$.

\medskip\noindent
\textbf{$\chi^2$-fields.} Recall that for a normalized GRF $\field$, we have $\field(t) \sim \mathcal{N}(0,1)$ for each $t  \in \Z^d$. In some cases, this assumption can be too restrictive, \eg, the distribution of (normalized) Gaussian random numbers is always symmetrical. As mentioned above, GRFs can be used as building blocks for more complex models, \eg,  $\chi^2$-fields.
Therefore, for some integer $n>0$, let $\field_1, \dots, \field_n$ be  independent and identically distributed  normalized GRFs. Then, the random field $\chiOne$ given by
\begin{equation}
	\chiOne = \sum_{i=1}^n \field_i^2
\end{equation}
is called a $\chi^2$-field with $n$ degrees of freedom. Moreover, we can model correlated $\chi^2$-fields by considering correlated GRFs. Therefore, let $(X_1,Y_1),\dots,(X_n,Y_n)$ be pairs of independent copies of $(\fieldOneCorrelated,\fieldTwoCorrelated)$, where the GRFs $\fieldOneCorrelated$ and $\fieldTwoCorrelated$ have been introduced in  Eq.~(\ref{eq:correlatedFields}). Then, the random fields $\chiOne$ and $\chiTwo$
given by
\begin{equation}\label{eq:chiSquaredFields}
	\chiOne = \sum_{i=1}^n \field_i^2 \quad \text{and} \quad  \chiTwo = \sum_{i=1}^n Y_i^2 
\end{equation}
are, in general, non-independent $\chi^2$-fields with $n$ degrees of freedom.

\subsubsection{Excursion sets}\label{sec:excursionSets}
The random field models considered in Section~\ref{sec:random_fields} themselves are not yet suitable for modeling morphologies (\eg, the phases of ASSB cathodes depicted in image data). Still, random fields are useful building blocks for modeling 3D morphologies, by considering excursion sets of random fields (sometimes  also referred to as level sets  \cite{adler2009random}). More precisely, for some ``level'' $\lambda \in \R$, the corresponding (random) excursion set $\Xi\subset\Z^d$ of a random field $\field=\{X(t),t\in\Z^d\}$ is given by
$\Xi=\{t \in \Z^d \colon \field(t) \geq \lambda\}$.
However, due to the deployment of convolutional neural networks in Section~\ref{sec:gan} which are supposed to receive these random sets as input, we will instead identify them as mappings $\Xi \colon \Z^d \to \{0,1\}$ that are given by
\begin{equation}\label{eq:excursion}
	\Xi(t) = \left\{
	\begin{array}{ll}
		1, & \text{if } \field(t) \geq \lambda,\\
		0, & \text{else,}
	\end{array}
	\right.
\end{equation}
for each $t\in \Z^d$. 
Such a mapping $\Xi$ can be considered to be 
a stochastic 3D model for the morphology of two-phase nano/microstructures. More precisely, for $d=3$, points $t\in \Z^3$ such that $\Xi(t)=1$ can be associated with one phase, whereas $1-\Xi$ can  describe the other phase.

Recall that the  cathode materials of ASSBs considered in the present paper are comprised of three phases. A common way to model the microstructure of such three-phase materials is to consider excursion sets of two (possibly correlated) random fields. More precisely, let $\fieldOne$ and $\fieldTwo$ be two random fields and $\lx,\ly \in \R$ be real-valued thresholds. Then, the first phase  $\Xi_1$ can be modeled as in Eq.~(\ref{eq:excursion}) by
\begin{equation}
	\Xi_1(t) = \left\{
	\begin{array}{ll}
		1, & \text{if }  \fieldOne(t) \geq \lx,\\
		0, & \text{else,}
	\end{array}
	\right.
\end{equation}
for each $t\in\Z^d$. 
The second phase  $\Xi_2$ can be modeled as the excursion set on the ``complement'' of $\Xi_1$ by
\begin{equation}
	\Xi_2(t) = \left\{
	\begin{array}{ll}
		1, & \text{if } \Xi_1(t)=0 \text{ and } \fieldTwo(t) \geq \ly,\\
		0, & \text{else,}
	\end{array}
	\right.
\end{equation}
for each $t\in \Z^d$.
Finally, the third phase is modeled by $\Xi_3= 1 - \Xi_1 - \Xi_2$.

\subsubsection{Isotropic three-phase model for ASSB cathodes}\label{sec:cathodeModel}
With  random field models and  corresponding excursion sets at hand (see Sections~\ref{sec:random_fields} and \ref{sec:excursionSets}), we can derive  models for the microstructure of three-phase materials. Recall that we will first derive an isotropic model for ASSB cathodes. Later, in Section \ref{sec:Cylindircallyisotropic}, we modify this model to capture the anisotropy (\ie, cylindrical isotropy) observed in experimentally measured 3D image data. For modeling three-phase microstructures isotropically with sufficient flexibility we consider the random fields $\fieldOne$, $\fieldTwo$, $\chiOne$, $\chiTwo$,  
where $\fieldOne$, $\fieldTwo$ are independent GRFs  which are independent of the correlated $\chi^2$-fields $\chiOne$, $\chiTwo$, see Eqs.~(\ref{eq:correlatedFields}) and (\ref{eq:chiSquaredFields}).
Now, we define the first and second phases $\Xi_1$, $\Xi_2$ by
\begin{equation}\label{eq:XiOneXiTwo}
	\Xi_1(t) = \left\{
	\begin{array}{ll}
		1, & \text{if }  \chiOne(t) + \sigma_\fieldOne \fieldOne(t) \geq \lx,\\
		0, & \text{else,}
	\end{array}
	\right.
	\quad \text{and} \quad
	\Xi_2(t) = \left\{
	\begin{array}{ll}
		1, & \text{if } \Xi_1(t)=0 \text{ and }  \chiTwo(t) + \sigma_\fieldTwo \fieldTwo(t) \geq \ly,\\
		0, & \text{else,}
	\end{array}
	\right.
\end{equation}
for each $t\in \Z^d$, where $\lx,\ly \in \R$  are some thresholds and  $\sigma_\fieldOne,\sigma_\fieldTwo > 0$ are scaling parameters that control the variance of the GRFs $\sigma_\fieldOne \fieldOne,\sigma_\fieldTwo \fieldTwo$, \ie, $\sigma_\fieldOne \fieldOne$ is a stationary isotropic GRF with $\mathrm{Var}\!\left( \sigma_\fieldOne \fieldOne(o) \right) = \sigma_\fieldOne^2 \mathrm{Var}\!\left( \fieldOne(o) \right)  = \sigma_\fieldOne^2$.\footnote{This scaling is omitted for the GRFs from which the $\chi^2$-fields $\chiOne,\chiTwo$ are constructed, since $\chi^2$-distributed random variables are, by definition, the squared sum of standard normally distributed random variables.
}

In the context of ASSB cathodes, $\Xi_1$ corresponds to the solid electrolyte, while $\Xi_2$ represents the active material phase.
Then, the third (pore) phase $\Xi_3$ is given by
$\Xi_3= 1 - \Xi_1 - \Xi_2$.
This leads to the  isotropic cathode model  $\cathodeModel \colon \Z^d \times \{1,2,3\} \to \{0,1\}$, which is given by
\begin{equation}\label{eq:cathodeModel}
	\cathodeModel(t,i)= \Xi_i(t)
	%
\end{equation}
for each $t\in \Z^d$ and $i\in \{1,2,3\}$. Clearly, it holds that $\cathodeModel(t,1)+\cathodeModel(t,2)+\cathodeModel(t,3)=1$ for each $t\in \Z^d$, \ie, $\cathodeModel(t,i)=1$ if and only if the grid point $t$ is associated with the $i$-th phase. 

Note that the model $\cathodeModel$ introduced in Eq.~\eqref{eq:cathodeModel}  is parametric. In particular,
the underlying  GRFs $\sigma_\fieldOne \fieldOne, \sigma_\fieldTwo \fieldTwo$ are parameterized by the parameters of their corresponding kernels and their scaling factors $\sigma_\fieldOne, \sigma_\fieldTwo$, see Eq.~(\ref{eq:isotropicKernel}).
Furthermore, recall that the $\chi^2$-fields $\chiOne$ and $\chiTwo$ are defined
by means of three independent GRFs $\widetilde{X},\widetilde{Y},\widetilde{Z}$, see Eqs.~(\ref{eq:correlatedFields}) and (\ref{eq:chiSquaredFields}).
Thus, the  $\chi^2$-fields $\chiOne$ and $\chiTwo$ are parameterized by $\gamma$ and the parameters of the kernels associated with $\widetilde{X},\widetilde{Y},\widetilde{Z}$. However,  the degree of freedom $n$ is considered to be fixed from now on. In particular, we choose $n=2$, motivated by \cite{NEUMANN2023112394}.
The remaining two parameters are the thresholds $\lx,\ly$ considered in Eq.~\eqref{eq:XiOneXiTwo}. The entire parameter vector of $\Xi$ will be denoted by $\theta =(\theta_1, \ldots, \theta_p) \in \Theta$, where $\Theta\subset\R^{p}$ is the set of all admissible parameter vectors with $p>0$ being the number of parameters. Furthermore, we will write $\Xi_\theta$ instead of $\Xi$ if we want to emphasize that $\Xi$ is characterized by the parameter vector $\theta\in\Theta$.  
Note that the number  of parameters of $\Xi$ is given by $p= 5 \cdot L +5$ since the isotropic cathode model $\Xi$ is based on five random fields, each of which is described by $L$ parameters of its kernel \footnote{The unnormalized kernels introduced in Eq.~\eqref{eq:kernelNonNormalized} have $L+1$ parameters, but due to the normalization performed in Eq.~\eqref{eq:isotropicKernel} the considered kernels have effectively $L$ parameters.
}. In addition, $\Xi$ is characterized by five further parameters, namely, $\gamma,\sigma_\fieldOne,\sigma_\fieldTwo,\lx,\ly$. As we set the  number $L$ of kernel parameters equal to 100 (\ie, kernels have a ``length'' of 201 and are centered at the origin $o\in\Z^d$), the cathode model $\Xi$ has in total $p=505$ parameters.
Later, in Section~\ref{sec:lpModels}, this relatively high-parametric cathode model  will be utilized to derive low-parametric cathode models, which are more suitable for the  purpose of virtual materials testing, as outlined in Section~\ref{sec:introduction}.

\subsection{Model calibration}\label{sec:modelCalibration}
In this section, we describe methods for calibrating the parametric cathode model $\cathodeModel_\theta$ introduced in Eq.~\eqref{eq:cathodeModel} to data, \ie, methods for determining a parameter vector $\fittedParam\in\Theta$ such that  $\fittedCathodeModel$ ``fits best'' to data. In \cite{NEUMANN2023112394,neumann2019pluri} similar excursion set models have been calibrated to image data, where  volume fractions and so-called two-point coverage probability functions have been used as geometrical descriptors to be matched. Therefore, we first introduce these quantities and explain how they can be estimated from data.

\subsubsection{Estimation of volume fractions and two-point coverage probability functions}
Let $\cathodeModel$ denote a (stationary and isotropic) random set model for three-phase morphologies, \eg, like the ASSB cathode model stated in Section~\ref{sec:cathodeModel}. Then, the volume fraction $\varepsilon_i$ of the $i$-th phase is equal to the probability that the origin $o\in\Z^d$ belongs to  the $i$-th phase, \ie, 
$\varepsilon_i=\probability{\Xi(o,i)=1}$,
for each $i \in \{1,2,3\}$.
Note that, due to the stationarity of $\cathodeModel$, the value of $\varepsilon_i$ does not depend on the chosen reference point $o\in\Z^d$, \ie, we have $\varepsilon_i= \probability{\Xi(t,i)=1}$ for any $t\in \Z^d$ and $i \in \{1,2,3\}$. 

Typically, we do not observe  realizations of the random set model $\Xi$ on the entire grid $\Z^d$, but on a bounded (typically cuboidal) sampling window $\samplingWindow \subset \Z^d$ instead. Then, due to stationarity of $\Xi$,  
\begin{equation}\label{eq:volumeFraction}
	\widehat{\varepsilon}_i = \frac{1}{\#\samplingWindow} \sum_{t\in \samplingWindow} \cathodeModel(t,i)
\end{equation}
is an unbiased estimator   for $\varepsilon_i$, for each phase $i\in\{1,2,3\}$, where $\#W$ denotes cardinality of the set $W$.

Furthermore, we consider a generalization of the volume fraction, namely, so-called two-point coverage probability functions. 
More precisely, for the $i$-th and $j$-th phase with $i,j\in\{1,2,3\}$, the two-point coverage probability function $C_{ij} \colon \{|t| \colon t \in \Z^d \} \to [0,1]$ is defined by
$C_{ij}(h) = \probability{\cathodeModel(o,i)=1, \cathodeModel(t_h,j)=1}$,
where $h\in\{|t| \colon t \in \Z^d \} $ and $t_h \in \Z^d$ is some point with $|t_h|=h$, \ie, $C_{ij}(h)$ is the probability that the origin $o$ is associated with the $i$-th phase, while a point $t_h$ with distance $h$ to the origin is associated with the $j$-th phase.

The estimation of $C_{ij}(h)$ from data is not as straightforward as for volume fractions. First, consider the function $c_{ij}\colon \Z^d \to [0,1]$, which is given by
\begin{equation}\label{eq:tpcpEstimator}
	c_{ij}(t) =  \left\{
	\begin{array}{ll}\displaystyle\frac{1}{c(t)} \sum_{s\in W, s+t \in W} \cathodeModel(s,i) \, \cathodeModel(s+t,j), &\mbox{if $ c(t)>0$.}\\
		0, & \mbox{else},
	\end{array}
	\right.    
\end{equation}
for each $t\in \Z^d$, where $c(t)= \#\{ s\in W \colon s+t \in W \}$. Then, for each $t\in \Z^d$ such that $c(t)>0$,  $c_{ij}(t)$ is an  unbiased estimator for $C_{ij}(h)$ with $h=|t|$. Note that for some distances  $h\in\{|t| \colon t \in \Z^d \} $, there might be several points $t\in \Z^d$ with $|t|\approx h$. Consequently, there might be multiple  estimators $c_{ij}(t)$ for $C_{ij}(h)$.
Therefore, we will use kernel regression to combine all these estimators. In this manner we obtain an estimator $\widehat{C}_{ij}(h)$ for arbitrary distances $h\ge 0$ such that $h=|t|$ and $c(t)>0$ for some $t\in\Z^d$.
With the notation $\mathcal{A}= 
\left\{
\left(|t|,c_{ij}(t) \right) \colon t\in \Z^d, c(t) >0
\right\},$
we can define the quantity $\widehat{C}_{ij}(h)$ even for an arbitrary distance $h\ge 0$ by putting
\begin{equation}\label{eq:tpcp}
	\widehat{C}_{ij}(h) = 
	\frac{
		\sum_{(x,y)\in \mathcal{A}} K(\frac{h-x}{b})\,y
	}{
		\sum_{(x,y)\in \mathcal{A}} K(\frac{h-x}{b})
	}\,,
\end{equation}
where $K:\R^d\to(0,\infty)$ denotes the Gaussian kernel and $b>0$ the bandwidth, which we set to  $b=0.5$.

Until now, the estimation of two-point coverage probability functions has been explained for single realizations of the random set model $\Xi$. Similarly, this approach can be deployed for multiple realizations. In particular, for $d=2$, Eq.~\eqref{eq:tpcp} 
can be utilized to compute two-point coverage probability function from from the experimentally measured 2D images $S_1, \dots, S_\numSlices \colon \samplingWindowTwo \times \{1,2,3\} \to \{0,1\}$  given in Eq.~\eqref{eq:slices}. More precisely,
by substituting $\cathodeModel$ with $S_k$ in Eqs.~\eqref{eq:tpcpEstimator} and \eqref{eq:tpcp}, we obtain the two-point coverage probability function associated with $S_k$, which we denote by $\widehat{C}_{ij}^{(k)}$ for each $k\in \{1,\ldots,\numSlices\}$.
Then, a single two-point probability function $\widehat{C}_{ij}^\mathrm{data}:[0,\infty)\to[0,1]$ associated with the experimentally measured 2D images is acquired by pointwise averaging, \ie, for each $h\geq 0$ we put
\begin{equation}\label{est.cov.dat}
	\widehat{C}_{ij}^\mathrm{data}(h)= \frac{1}{\numSlices} \sum_{k=1}^\numSlices \widehat{C}_{ij}^{(k)}(h).
\end{equation}


\subsubsection{Fitting approximative excursion set models}\label{sec:tpp_fitting}
Note that, for some random field models,  there are explicit analytical formulas which link the two-point coverage probabilities $C_{ij}(h)$ of excursion sets with the covariance functions of the underlying random fields and their thresholds, \eg, for excursion sets of correlated GRFs \cite{neumann2019pluri}. For such models,  this allows for the direct calibration of model parameters to image data. However, for more complex models, like the isotropic cathode model $\cathodeModelParametric$ introduced in Section~\ref{sec:cathodeModel}, to our knowledge  such formulas do not exist. 
Furthermore, even if such formulas existed, minor changes to the model would immediately bring about the need to derive updated formulas. Therefore, in this section we describe a computational method that enables a computer-assisted calibration of $\cathodeModelParametric$ and similar excursion set models to image data. 


For each pair $(i,j) \in I=\{(x,y) \colon x,y\in \{1,2,3\} \mbox{ with $x\leq y$}\}$, 
let $\widehat{C}_{ij}^\mathrm{data}:[0,\infty)\to[0,1]$ denote the two-point coverage probability function estimated from image data by means of Eq.~\eqref{est.cov.dat}. In addition, for each parameter vector $\theta \in \Theta$ we denote the two-point coverage probability functions of $\cathodeModelParametric$ by ${C}_{ij,\theta}:\{|t|:t\in\Z^d\}\to[0,1]$ for each pair $(i,j) \in I$. In order to quantify the discrepancy of the two-point coverage probability functions associated with the model $\cathodeModelParametric$ and those estimated from image data, we introduce the loss function $\mathrm{loss} \colon \Theta \to [0,\infty)$ given by
\begin{equation}\label{eq:lossFunction}
	\mathrm{loss}(\theta)=\sum_{(i,j)\in I} \sum_{h=0}^{h_\mathrm{max}} 
	\left(
	\widehat{C}_{ij}^\mathrm{data}(h)-{C}_{ij,\theta}(h)
	\right)^2,
\end{equation}
for each $\theta \in \Theta$, where we set the maximal distance considered to $h_\mathrm{max}=100$. 
Then, we can define the ``optimal'' parameter $\fittedParam\in\Theta$ by considering the minimization problem
\begin{equation}\label{eq:minimizationProblem}
	\fittedParam= \argmin_{\theta \in \Theta} \mathrm{loss}(\theta).
\end{equation}
Note that no analytical formulas exist for the direct computation of the minimum in Eq.~\eqref{eq:minimizationProblem}. Thus, we have to deploy iterative numerical methods which provide an approximate solution rather than an exact minimum. Typically, numerical minimization methods require repeated evaluations of the term on the right-hand side of Eq.~\eqref{eq:minimizationProblem}, \ie, of $\mathrm{loss}(\theta)$ for different parameter vectors $\theta \in \Theta$. 
However, this necessitates the computation of the values ${C}_{ij,\theta}(h)$, which cannot be determined explicitly due to the complexity of the underlying model $\cathodeModelParametric$. 
To remedy this problem,  we compute approximations of ${C}_{ij,\theta}(h)$ by means of Monte-Carlo simulation \cite{kroese2013handbook}. 
More precisely,
we  simulate a realization $\xi_\theta$ of the model $\cathodeModelParametric$ on some sufficiently large sampling window $\samplingWindow$ followed by the computation of the estimate $\widehat{C}_{ij}(h)$ by means of Eq.~(\ref{eq:tpcp}). From here on, we denote such estimates computed for model realizations $\xi_\theta$ of $\cathodeModelParametric$ by $\widehat{C}_{ij,\theta}(h)$. By substituting ${C}_{ij,\theta}(h)$ with its approximation $\widehat{C}_{ij,\theta}(h)$ in Eq.~\eqref{eq:lossFunction}, we obtain a loss function $\widehat{ \mathrm{loss}}\colon \Theta \to [0,\infty)$ given by
\begin{equation}\label{eq:lossFunctionApprox}
	\widehat{\mathrm{loss}}(\theta)=\sum_{(i,j)\in I} \sum_{h=0}^{h_\mathrm{max}} 
	\left(
	\widehat{C}_{ij}^\mathrm{data}(h)-\widehat{C}_{ij,\theta}(h)
	\right)^2,
\end{equation} 
which can be numerically evaluated for each $\theta \in \Theta$ and approximates the original loss function given in Eq.~\eqref{eq:lossFunction}.\footnote{Note that due to approach based on Monte-Carlo simulation, the values 
	of $\widehat{C}_{ij,\theta}(h)$ depend on model realizations of $\cathodeModelParametric$, \ie, the values $\widehat{C}_{ij,\theta}(h)$ and consequently the values of $\widehat{\mathrm{loss}}(\theta)$ are random approximations of ${C}_{ij,\theta}(h)$.
} 

\begin{figure}[H]
	\centering
	\begin{tikzpicture}
		\node[draw, rounded corners=5pt,thick,align=center,text width=\imageLength] (params)  {
			\textbf{parameter vector}
			$\theta$\\
			(kernel parameters \&\\
			thresholds)
		};
		
		\node[right=0.75\imageLength of params.east, anchor=west,draw, rounded corners=5pt,thick,align=center,text width=0.8\imageLength] (kernels)  {
			\textbf{kernels}\\ $\kernel_\alpha$
		};
		
		\node[above=0.05\imageLength of kernels.north, anchor=south,draw, rounded corners=5pt,thick,align=center,text width=0.8\imageLength] (noise)  {
			\textbf{noise} $\noise$
		};
		
		\node[right=0.75\imageLength of kernels.east, anchor=west,draw, rounded corners=5pt,thick,align=center,text width=1.75\imageLength] (GRFs)  {
			\textbf{independent GRFs}\\ 
			$\fieldOne,\fieldTwo,\widetilde{X},\widetilde{Y},\widetilde{Z}$
		};
		
		\node[right=0.75\imageLength of GRFs.east, anchor=west,draw, rounded corners=5pt,thick,align=center,text width=1.5\imageLength] (corGRFs)  {
			\textbf{correlated GRFs}\\ 
			$\fieldOne,\fieldTwo,\fieldOneCorrelated,\fieldTwoCorrelated$
		};

		\node[below=0.3\imageLength of corGRFs.south, anchor=north,draw, rounded corners=5pt,thick,align=center,text width=1.5\imageLength] (chiSquared)  {
			\textbf{correlated $\chi^2$-fields and GRFs}\\ 
			$\fieldOne,\fieldTwo,\chiOne,\chiTwo$
		};

		\node[left=\imageLength of chiSquared.west, anchor=east,draw, rounded corners=5pt,thick,align=center,text width=1.3\imageLength] (excursion)  {
			\textbf{excursion set model} $\cathodeModelParametric$
		};
		
		\node[below=0.3\imageLength of excursion.south, anchor=north,draw, rounded corners=5pt,thick,align=center,text width=1.3\imageLength] (excursionApprox)  {
			\textbf{approximative model} $\cathodeModelParametricApprox$
		};

		\node[left=\imageLength of excursion.west, anchor=east,draw, rounded corners=5pt,thick,align=center,text width=2.5\imageLength] (tpcp)  {
			\textbf{two-point coverage \\ probability functions} $\widehat{C}_{ij,\theta}$
		};

		\node[below=0.3\imageLength of tpcp.south, anchor=north,draw, rounded corners=5pt,thick,align=center,text width=2.5\imageLength,yshift=0.15em] (tpcpApprox)  {
			\textbf{two-point coverage \\ probability functions} $\widetilde{C}_{ij,\theta}$
		};

		\draw[vecArrow] ([xshift=0.05\imageLength]params.east) to node[below,align=center] {
			\small
			Eqs. (\ref{eq:kernelNonNormalized}) \\ \small and (\ref{eq:isotropicKernel})
		}	([xshift=-.05\imageLength]kernels.west);

		\draw[vecArrow] ([xshift=0.05\imageLength]kernels.east) to node[below,align=center] {
			\small
			Eq. (\ref{eq:moving})
		}	([xshift=-.05\imageLength]GRFs.west);
		
		\draw[vecArrow] ([xshift=0.05\imageLength]noise.east) -|	([xshift=-.25\imageLength,yshift=0.01\imageLength]GRFs.west);

		\draw[vecArrow] ([xshift=0.05\imageLength]GRFs.east) to node[below,align=center] {
			\small
			Eq. (\ref{eq:correlatedFields})
		}	([xshift=-.05\imageLength]corGRFs.west);
		
		\draw[vecArrow] ([yshift=-0.05\imageLength]corGRFs.south) to node[left,align=center] {
			\small
			Eq. (\ref{eq:chiSquaredFields})
		}	([yshift=.05\imageLength]chiSquared.north);

		\draw[redvecArrow] ([xshift=-0.05\imageLength]chiSquared.west) to node[below,align=center] {
			\small
			\textcolor{black}{Eqs.} (\ref{eq:XiOneXiTwo})\textcolor{black}{}\\
			\small \textcolor{black}{
				and (\ref{eq:cathodeModel})}
		}	([xshift=.05\imageLength]excursion.east);

		\draw[vecArrow] ([xshift=-0.05\imageLength]excursion.west) to node[below,align=center] {
			\small
			Eqs. \\(\ref{eq:tpcpEstimator}) and (\ref{eq:tpcp})
		}	([xshift=.05\imageLength]tpcp.east);

		\draw[vecArrow] ([yshift=-0.05\imageLength]chiSquared.south) |-
		node[below,align=center,xshift=-\imageLength]{Eq. (\ref{eq:XiOneXiTwoAapprox})}	([xshift=.05\imageLength]excursionApprox.east);

		\draw[vecArrow] ([xshift=-0.05\imageLength]excursionApprox.west) to node[below,align=center] {
			\small
			Eqs. \\(\ref{eq:tpcpEstimator}) and (\ref{eq:tpcp})
		}	([xshift=.05\imageLength]tpcpApprox.east);
		
	\end{tikzpicture}
	\caption{Computational scheme for mapping the parameter vector $\theta$ onto a realization $\cathodeModelParametric$ and a corresponding estimate of the two-point coverage probability functions $\widehat{C}_{ij,\theta}$ (first and second rows). Arrows visualized in black indicate operations that are differentiable with respect to  $\theta$. The red arrow indicates a non-differentiable operation, \ie, the thresholding performed in Eq.~(\ref{eq:XiOneXiTwo}) for computing excursion sets is non differentiable. As an alternative a differentiable approximation is proposed (third row).}
	\label{fig:computationalScheme}
\end{figure}
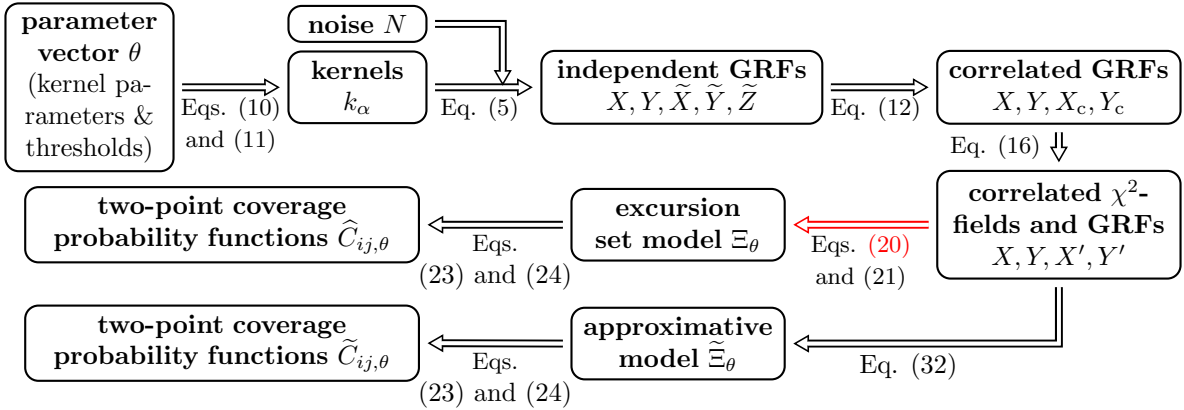

There are numerous numerical methods for minimizing functions, some of which utilize gradient information (\ie, gradient descent methods) to efficiently explore the parameter space  $\Theta$. In order to deploy gradient descent methods, we must be able to compute the gradient of $\widehat{\mathrm{loss}}$ which is given by
\begin{equation}
	\nabla_\theta \widehat{\mathrm{loss}}(\theta) = \left( \frac{\partial}{\partial \theta_1} \widehat{\mathrm{loss}}(\theta), \ldots, \frac{\partial}{\partial \theta_p} \widehat{\mathrm{loss}}(\theta) \right),
\end{equation}
where $\frac{\partial}{\partial \theta_k} \widehat{\mathrm{loss}}(\theta)$ denotes the partial derivatives of $\widehat{\mathrm{loss}}$ for each $k\in \{1,\ldots,p\}$. In particular, due to the chain rule of differentiation, this would necessitate the computation of the partial derivatives of $\widehat{C}_{ij,\theta}(h)$ and, even more critically, the partial derivatives of $\xi_\theta(x,i)$ for each $x \in W$ and $i\in \{1,2,3\}$. However, the function $\theta \mapsto \xi_\theta(x,i)$ is not differentiable, and thus the partial derivatives $\frac{\partial}{\partial \theta_k} \xi_\theta(x,i)$ do not exist, see Figure~\ref{fig:computationalScheme} (red arrow). This becomes evident when considering an alternative representation of $\cathodeModelParametric$ given by
\begin{equation}\label{eq:XiOneXiTwoAlternative}
	\cathodeModelParametric(t,1) = \heaviside(\chiOne(t) + \fieldOne(t) - \lx)
	\quad \text{and} \quad
	\cathodeModelParametric(t,2) = \heaviside(\chiTwo(t) + \fieldTwo(t) - \ly) \cdot (1 - \cathodeModelParametric(t,1)),
\end{equation}
for each $t \in \Z^d$, where $\heaviside= \mathbbm{1}_{[0,\infty)}$ denotes the  Heaviside step function which is not differentiable at $0$. Moreover, the derivatives of $\heaviside$ are zero everywhere else, which implies that, if the loss function $\widehat{\mathrm{loss}}$ is differentiable, the gradient $\nabla_\theta \widehat{\mathrm{loss}}(\theta)$ would be a null vector.
This phenomenon is subsumed under the vanishing gradient problem, which makes minimizing the loss function using gradient descent impossible.

We can remedy this issue by  substituting $\heaviside$ with a differentiable approximate with non-vanishing gradients, \eg, a sigmoid-type function. 
More precisely, the function $\widetilde{\heaviside}$ given by
\begin{equation}
	\widetilde{\heaviside}(r) = \mathrm{sigmoid}(\nu r) = \frac{1}{1+e^{-\nu r}},
\end{equation}
for each $r\in \R$, is differentiable and converges  to $\heaviside$ for $\nu \to \infty$ (we observed that the parameter choice of $\nu=10$ is  reasonable). 
Then, by substituting $\heaviside$ with 
$\widetilde{\heaviside}$ in Eq.~\eqref{eq:XiOneXiTwoAlternative}, 
we obtain an differentiable and approximative version $\widetilde{\Xi}_\theta\colon \Z^d \times \{1,2,3\} \to (0,1)$ of the model $\cathodeModel_\theta$ that is given by
\begin{equation}\label{eq:XiOneXiTwoAapprox}
	\widetilde{\Xi}_\theta(t,i) = 
	\left\{
	\begin{array}{ll}
		\widetilde{\heaviside}(\chiOne(t) + \fieldOne(t) - \lx), & \text{if } i=1,\\
		\widetilde{\heaviside}(\chiTwo(t) + \fieldTwo(t) - \ly) \cdot (1 - \widetilde{\Xi}_\theta(t,1)), & \text{if } i=2,\\
		1- \widetilde{\Xi}_\theta(t,1) -\widetilde{\Xi}_\theta(t,2), & \text{if } i=3,
	\end{array}
	\right.
\end{equation}
for each $t\in \Z^d$ and $i\in\{1,2,3\}$.

Note that by substituting $\cathodeModel$ with $\cathodeModelParametricApprox$ in Eq.~(\ref{eq:tpcpEstimator}), followed by the computation performed in Eq.~(\ref{eq:tpcp}), we determine an approximative estimate of the two-point probability function $\widehat{C}_{ij,\theta}$ which we denote by $\widetilde{C}_{ij,\theta}$. The advantage of considering  $\widetilde{C}_{ij,\theta}$ instead of  $\widehat{C}_{ij,\theta}$ is that the former is differentiable and has non vanishing gradients with respect to the parameter vector $\theta$. Therefore, we can deploy gradient descent algorithms to calibrate the excursion set model $\cathodeModelParametric$, \ie, to solve the problem given in Eq.~(\ref{eq:minimizationProblem}), using the loss function
\begin{equation}\label{eq:lossFunctionApproxDiff}
	\widetilde{\mathrm{loss}}(\theta)=\sum_{(i,j)\in I} \sum_{h=0}^{h_\mathrm{max}} 
	\left(
	\widehat{C}_{ij}^\mathrm{data}(h)-\widetilde{C}_{ij,\theta}(h)
	\right)^2,
\end{equation}
which approximates the loss function given in Eq.~\eqref{eq:lossFunction} for each $\theta \in \Theta$.
In particular, we use a stochastic gradient descent approach to iteratively improve the model parameter $\theta$, see Algorithm~\ref{alg:training_procedure}.

We have implemented Algorithm~\ref{alg:training_procedure} in the Python package PyTorch \cite{NEURIPS2019_9015}, which enables us to exploit automatic differentiation, \ie, partial derivatives of loss functions with respect to model parameters are automatically derived by applying the chain rule. Moreover, in our implementation we deploy fast Fourier transformation to accelerate the computation of convolutions between kernels and noise \cite{abdallah2016morphological}---which is performed for simulating GRFs on the sampling window $\samplingWindow$, see Eq.~(\ref{eq:moving}). Similarly, note that the operation in Eq.~(\ref{eq:tpcpEstimator}) (which is necessary to estimate two-point coverage probability functions) can be formulated by means of convolutions. Consequently, the computation of two-point coverage probabilities can be accelerated by deploying fast Fourier transformations as well.

After training, we can generate realizations of the model $\cathodeModel_{\fittedParam}$, see Figure~\ref{fig:model_realization_tpp}a. By visual inspection the morphology of these realizations statistically deviate quite strongly from the training data shown in Figure~\ref{fig:anistropy}b. In particular the boundaries of phases seem to be too rough in model realizations. 
An explanation for this could be that Algorithm~\ref{alg:training_procedure} can be considered to be a computational approach of solving the underdetermined system of equations $\widetilde{C}_{ij}^\mathrm{data}={C}_{ij,\theta}$, where the model parameters $\theta$ are the unknown variables.

\begin{algorithm}[H]
	\caption{Training approach using two-point probability functions (\textbf{Inputs:} 
		$\samplingWindow$: sampling window;
		$\widehat{C}_{ij}^\mathrm{data}$:
		two-point probability functions;  $n_\mathrm{epoch}$: number of epochs; $n_\mathrm{steps}$: steps  per epoch. \textbf{Output:} $\widehat{\theta}$: fitted parameter vector.)}\label{alg:training_procedure}
	\begin{algorithmic}[1]
		\Procedure{TrainTPPF}{$\widehat{C}_{ij}^\mathrm{data}, W, n_\mathrm{epochs}, n_\mathrm{steps}$}
		\State $\lr \gets 0.0001$ \Comment{Learning rate}
		\State $\bs \gets 32$  \Comment{Batch size}
		\State initialize $\theta$ by simulating Gaussian noise \Comment{Parameter vector}
		\For{$e \text{ in } 1,\ldots,n_\mathrm{epoch}$} \Comment{Epochs}
		\For {step in $ 1,\ldots,n_\mathrm{steps}$}\Comment{Model training steps}
		\State Generate $\widetilde{\xi}_1,\dots,\widetilde{\xi}_\bs$ i.i.d. copies of 		 $\widetilde{\cathodeModel}_\theta$ on sampling window $\samplingWindow$ \Comment{Model realizations}
		\State From $\widetilde{\xi}_\ell$ compute $\widetilde{C}_{ij,\theta}^{(\ell)}$
		for each $(i,j) \in I$ and $\ell =1,\dots,\bs$ \Comment{ Eqs.~(\ref{eq:tpcpEstimator}) and ~(\ref{eq:tpcp})}
		\State 
		$\widetilde{\mathrm{loss}}(\theta)\gets \frac{1}{\bs} \sum_{\ell=1}^{\bs} \sum_{(i,j)\in I} \sum_{h=0}^{h_\mathrm{max}} 
		\left(
		\widetilde{C}_{ij}^\mathrm{data}(h)-\widetilde{C}_{ij,\theta}^{(\ell)}
		\right)^2$ \Comment{Approximate loss in Eq.~(\ref{eq:lossFunctionApproxDiff})} 
		\State Update learning rate $\lr$ according to Adam algorithm \Comment{see \cite{Kingma2014}}
		\State $\theta \gets \theta - \lr \cdot \, \nabla_\theta \widetilde{\mathrm{loss}}(\theta)$ 
		\EndFor
		\EndFor
		\State $\widehat{\theta} \gets \theta$
		\State \Return $\widehat{\theta}$
		\EndProcedure
	\end{algorithmic} 
\end{algorithm}


\subsubsection{Generative adversarial framework for fitting excursion sets}\label{sec:gan}

\begin{figure}[ht]
	\centering
	\begin{subfigure}[b]{0.255\textwidth}
		\centering
		\includegraphics[width=\textwidth]{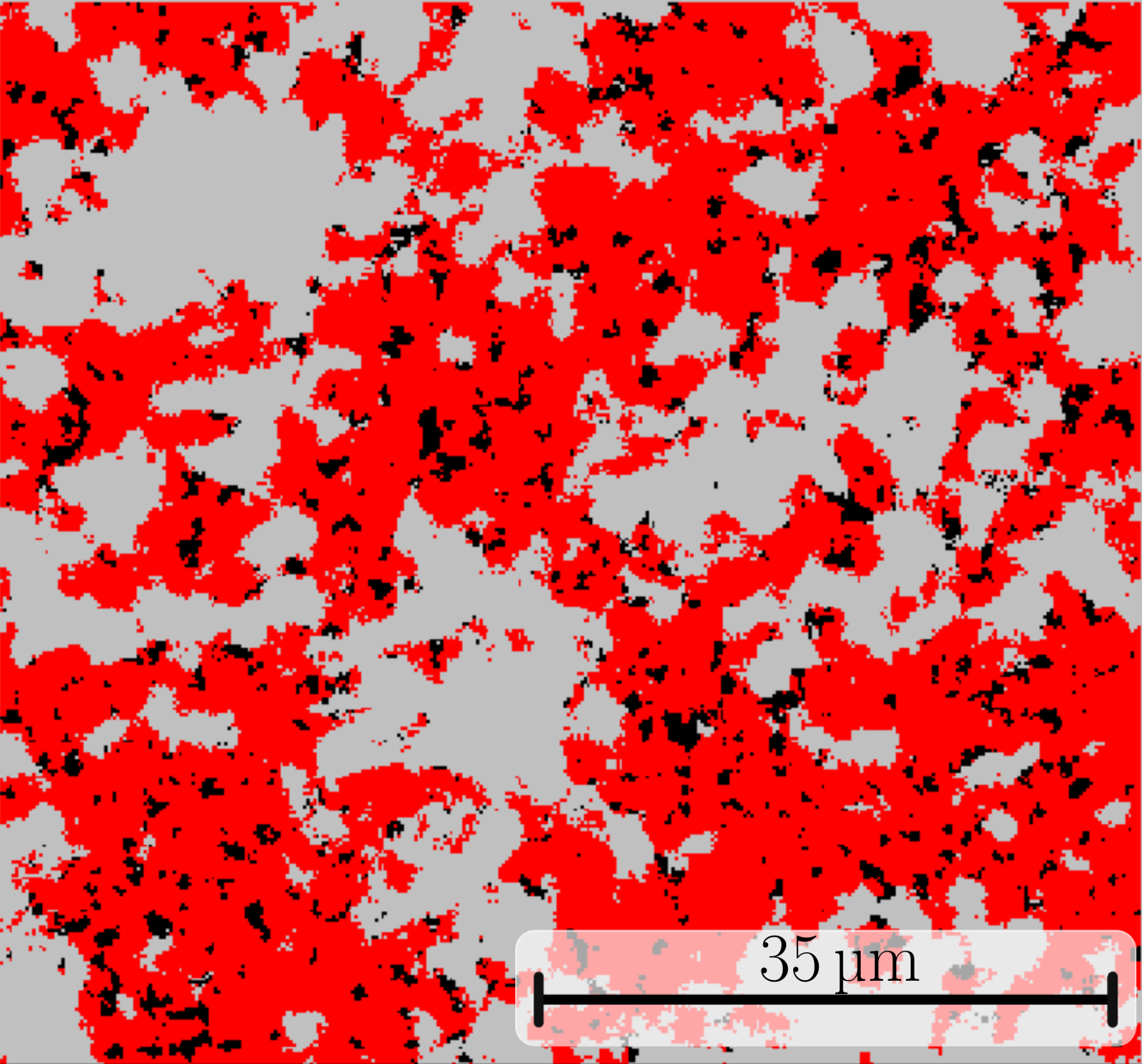}
		\caption{}		
	\end{subfigure}
	\hspace{1cm}
	\begin{subfigure}[b]{0.25\textwidth}
		\centering
		\includegraphics[width=\textwidth]{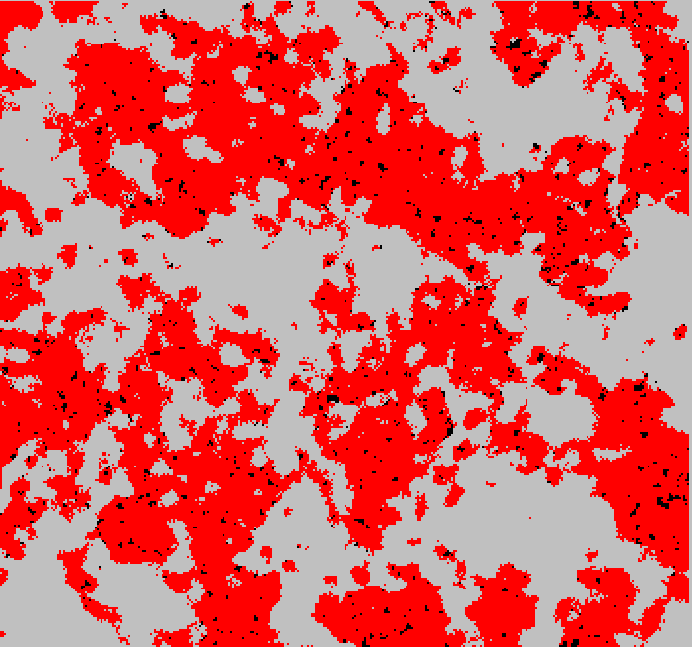}
		\caption{}		
	\end{subfigure}
	\hspace{1cm}
	\begin{subfigure}[b]{0.25\textwidth}
		\centering
		\includegraphics[width=\textwidth]{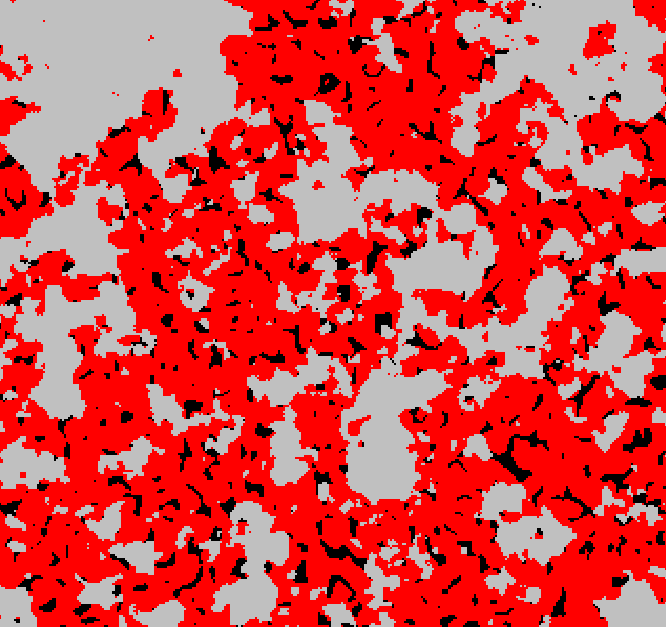}
		\caption{}		
	\end{subfigure}
	\caption{2D model realizations of a model trained with Algorithm \ref{alg:training_procedure} (a), a model trained with Algorithm \ref{alg:training_procedure_gan} (b) and a model trained with Algorithm \ref{alg:training_procedure_gan_combined} (c). The pore space, active material and the solid electrolyte are represented by black, red and gray color, respectively.  All figures use the same length scale.}
	\label{fig:model_realization_tpp}
\end{figure}

\begin{figure}[H]
	\centering
	\includegraphics[width=0.6\textwidth]{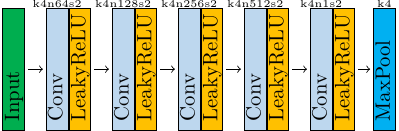}
	\caption{Visualization of the discriminator's network architecture. The $\alpha$ parameter of the LeakyReLU layers is set to 0.2, see \cite{Maas2013RectifierNI} for further details on LeakyReLu layers.
		The labels above convolutional layers (Conv) indicate the kernel size (k), the number of feature maps (n) and the stride (s), see \cite{Goodfellow2016} for more details on the deployed layers and their parameters. For example, the label k4n64s2 indicates a convolutional layer with a kernel size  of 9, 64 feature maps and a stride of 1.
	}
	\label{fig:discriminator}
\end{figure}

In the previous section a method has been described for the calibration of the parameters $\theta$ of the  model $\cathodeModel_{\theta}$ by minimizing the discrepancy between two-point coverage probability functions estimated from model realizations and data, \ie, user-defined statistics have been used for model calibration which might be insufficient for deriving a digital twin of the data, see Figure~\ref{fig:model_realization_tpp}a.
Note that there are data driven methods that can learn further statistics for distinguishing between model realizations and image data.
Therefore, let $\cathodeModelCutout$ be a random cutout  taken from the image data $S_1, \dots, S_\numSlices$ such that the size of the sampling window of $\cathodeModelCutout$ coincides with the sampling window $\samplingWindow$, \ie, we have that $\cathodeModelCutout \colon \samplingWindow \times \{1,2,3\} \to \{0,1\}$. Now assume that we have some function $\discriminator$ with values in the interval $[0,1]$ called discriminator that can distinguish between $\cathodeModelCutout$ and realizations of the model $\cathodeModel_{\theta}$ restricted on $\samplingWindow$, where we denote the latter by $\cathodeModel_{\theta}\vert_\samplingWindow$ from here on. Roughly speaking, for a ``bad choice'' of the model parameter $\theta$, we have $\discriminator(\cathodeModelCutout) \approx 1$ and  $\discriminator(\cathodeModel_{\theta}\vert_\samplingWindow) \approx 0$, or alternatively 
$\discriminator( \widetilde{\cathodeModel}_{\theta}\vert_\samplingWindow) \approx 0$ when considering the approximative version $\widetilde{\cathodeModel}_{\theta}$ of ${\cathodeModel}_{\theta}$ instead. Then, the model parameter $\theta$ can be optimized with the goal to make $\widetilde{\cathodeModel}_{\theta}$ indistinguishable from $\cathodeModelCutout$ with respect to the discriminator $\discriminator$, \eg, by solving the following minimization problem
\begin{equation}\label{eq:gen_loss}
	\min_{\theta \in \Theta} \mathbb{E}\! \left[ (1-\discriminator(\widetilde{\cathodeModel}_{\theta} \vert_\samplingWindow))^2 \right].
\end{equation}
Typically, the discriminator $\discriminator$ is a convolutional neural network (CNN), \ie, it is parameterized by its layer weights, which we aggregate to $\discriminatorWeights \in \discriminatorWeightsSet$, where $\discriminatorWeightsSet$ denotes the set of all admissible weights. Thus, we denote a discriminator with weights $\discriminatorWeights$ by $\discriminator_\discriminatorWeights$. The network architecture considered in the present paper is visualized in Figure~\ref{fig:discriminator}.

Since the weights $\discriminatorWeights$ are typically randomly initialized it is to be expected that the discriminator $\discriminator_\discriminatorWeights$ is unable to distinguish between model realizations and image data---consequently, it has to be trained, \eg, by solving the maximization problem
\begin{equation}\label{eq:disc_loss}
	\max_{\discriminatorWeights \in \discriminatorWeightsSet} \mathbb{E}\! \left[ \discriminator_\discriminatorWeights(\cathodeModelCutout)^2 \right]+  \mathbb{E}\! \left[ (1-\discriminator_\discriminatorWeights(\widetilde{\cathodeModel}_{\theta} \vert_\samplingWindow))^2 \right].
\end{equation}
Since the argument $\theta$ within the objective function of the minimization problem (\ref{eq:gen_loss}) does not influence $\discriminator_\discriminatorWeights(\cathodeModelCutout)^2$, we can write the optimization problems  (\ref{eq:gen_loss}) and (\ref{eq:disc_loss}) as a minmax problem
\begin{equation}\label{eq:gan}
	\min_{\theta \in \Theta}\max_{\discriminatorWeights \in \discriminatorWeightsSet} \mathbb{E}\! \left[ \discriminator_\discriminatorWeights(\cathodeModelCutout)^2 \right]+  \mathbb{E}\! \left[ (1-\discriminator_\discriminatorWeights(\widetilde{\cathodeModel}_{\theta} \vert_\samplingWindow))^2 \right].
\end{equation}
Put simply, on one hand, the discriminator learns its own ``data-driven'' statistics (features of the convolutional layers), which enable it to distinguish between model realizations and image data. On the other hand, these statistics are then used for calibration of  $\widetilde{\cathodeModel}_{\theta}$.
Such minmax problems are often optimized alternatingly with iterative optimization methods, \ie,
see Algorithm~\ref{alg:training_procedure_gan} for our computational implementation.

\begin{algorithm}[H]
	\caption{Training approach based on generative adversarial framework (\textbf{Inputs:} $\samplingWindow$: sampling window; 
		$\cathodeModelCutout$: random cutouts  of image data  $S_1, \dots, S_\numSlices$ on $\samplingWindow$;
		$n_\mathrm{epoch}$: number of epochs; $n_\mathrm{steps}$: steps per epoch.
		\textbf{Output:} $\widehat{\theta}$: fitted parameter vector.)}\label{alg:training_procedure_gan}
	\begin{algorithmic}[1]
		\Procedure{Train}{$\cathodeModelCutout, \samplingWindow,n_\mathrm{epoch},n_\mathrm{steps}$}
		\State $\lr \gets 0.0001$ \Comment{Learning rate model}
		\State $\lr_\mathrm{D} \gets 0.0001$ \Comment{Learning rate discriminator}
		\State $\bs \gets 32$  \Comment{Batch size}
		\State initialize $\theta$ by simulating Gaussian noise \Comment{Parameter vector}
		\State initialize $\discriminatorWeights$ by simulating Gaussian noise \Comment{Weight vector}
		\For{$e \text{ in } 1,\ldots,n_\mathrm{epoch}$} \Comment{Epochs}
		
		\For {step in $ 1,\ldots,n_\mathrm{steps}$}\Comment{Model training steps}
		\State $\theta \gets \mathrm{GeneratorTrainingStep}(\theta,\discriminatorWeights,\samplingWindow,\bs,\lr)$ \Comment{Training step} 
		\EndFor
		
		\For {step in $ 1,\ldots,n_\mathrm{steps}$}\Comment{Discriminator training steps}
		\State $\discriminatorWeights \gets \mathrm{DiscriminatorTrainingStep}(\cathodeModelCutout,\theta,\discriminatorWeights,\samplingWindow,\bs,\lr_\mathrm{D})$ \Comment{Training step}
		\EndFor

		\EndFor
		\State $\widehat{\theta} \gets \theta$
		\State \Return $\widehat{\theta}$
		\EndProcedure
		\\
		
		\Procedure{GeneratorTrainingStep}{$\theta,\discriminatorWeights,\samplingWindow,\bs,\lr$}
		\State Generate $\widetilde{\xi}_1,\dots,\widetilde{\xi}_\bs$ i.i.d. copies of 		 $\widetilde{\cathodeModel}_\theta$ on sampling window $\samplingWindow$ \Comment{Model realizations}
		\State $\mathrm{loss}(\theta) \gets  \frac{1}{\bs} \sum_{\ell=1}^{\bs} (1-\discriminator_\discriminatorWeights(\widetilde{\xi}_{\ell} \vert_\samplingWindow))^2$	\Comment{Eq. (\ref{eq:gen_loss})}				
		\State Update learning rate $\lr$ according to Adam algorithm \Comment{see \cite{Kingma2014}}
		\State $\theta \gets \theta - \lr \cdot \, \nabla_\theta \loss(\theta)$ 
		\State \Return $\theta$
		\EndProcedure
		
		\\
		\Procedure{DiscriminatorTrainingStep}{$\cathodeModelCutout,\theta,\discriminatorWeights,\samplingWindow,\bs,\lr_\mathrm{D}$}
		\State Generate $\widetilde{\xi}_1,\dots,\widetilde{\xi}_\bs$ i.i.d. copies of 		 $\widetilde{\cathodeModel}_\theta$ on sampling window $\samplingWindow$ \Comment{Model realizations}
		\State Generate $\cathodeModelCutoutRealization_1,\dots,\cathodeModelCutoutRealization_\bs$ i.i.d. copies of 		 $\cathodeModelCutout$ \Comment{Sample data}
		\State $\mathrm{loss}_\mathrm{D}(\theta_\mathrm{D}) \gets - \frac{1}{\bs} \sum_{\ell=1}^{\bs} \discriminator_\discriminatorWeights(\cathodeModelCutoutRealization_\ell)^2+ (1-\discriminator_\discriminatorWeights(\widetilde{\xi}_{\ell} \vert_\samplingWindow))^2$	\Comment{Eq. (\ref{eq:disc_loss})}		
		\If{$\mathrm{loss}_\mathrm{D}(\theta_\mathrm{D})>0.4$}	 \Comment{Avoid discriminator overfitting \cite{fuchs2024generating}}	
		\State Update learning rate $\lr_\mathrm{D}$ according to Adam algorithm \Comment{see \cite{Kingma2014}}
		\State $\discriminatorWeights \gets \discriminatorWeights - \lr_\mathrm{D} \cdot \, \nabla_\discriminatorWeights \mathrm{loss}_\mathrm{D}(\theta_\mathrm{D})$ 
		\EndIf
		\State \Return $\discriminatorWeights$
		\EndProcedure
	\end{algorithmic} 
\end{algorithm}

Note that the discriminator architecture used in the present paper (see Figure~\ref{fig:discriminator}) receives 2D images as input, \ie, we consider sampling windows $\samplingWindow= \{1,\dots,201\}^2$. Thus, due to the assumption of isotropy, 2D image data suffices for the purpose of  calibrating a 3D model. A visualization of a model realization after training is shown in Figure~\ref{fig:model_realization_tpp}b. We believe that the statistical discrepancy  to the image data is caused by the slow convergence of Algorithm~\ref{alg:training_procedure_gan}. Typically, it is difficult to strike a balance between the learning rates of the discriminator and the model to be calibrated \cite{saad2024survey}.

\subsubsection{Combined fitting approach}\label{sec:combined}
In this section we combine the fitting approaches presented in Sections~\ref{sec:tpp_fitting} and \ref{sec:gan}. More precisely, for the purpose of model calibration, we consider a different loss function which uses both user-defined statistics (two point coverage probabilities) 
and features learned by a discriminator. 
In particular, we combine the optimization problems given by (\ref{eq:minimizationProblem}) and (\ref{eq:gan}) to obtain the optimization problem
\begin{equation}\label{eq:gancombined}
	\min_{\theta \in \Theta}\max_{\discriminatorWeights \in \discriminatorWeightsSet} \mathbb{E}\! \left[ \discriminator_\discriminatorWeights(\cathodeModelCutout)^2 \right]+  \mathbb{E}\! \left[ (1-\discriminator_\discriminatorWeights(\widetilde{\cathodeModel}_{\theta} \vert_\samplingWindow))^2 \right]
	+
	\gamma 
	\sum_{(i,j)\in I} \sum_{h=0}^{h_\mathrm{max}} 
	\left(
	\widehat{C}_{ij}^\mathrm{data}(h)-{C}_{ij,\theta}(h)
	\right)^2,
\end{equation}
where $\gamma>0$ is a weighting factor. The computational implementation for solving this minmax problem is given in Algorithm~\ref{alg:training_procedure_gan_combined}. A realization of the model $\cathodeModelParametric$ calibrated in this manner is visualized in Figure~\ref{fig:model_realization_tpp}c, which already indicates a relatively good match.

Note that for training purposes, we set the maximum number of epochs to $n_\text{epoch} = 5000.$ In order to improve the run times of the calibration with Algorithm~\ref{alg:training_procedure_gan_combined}, in practice we use an early stopping criterion that can terminate the calibration after 100 epochs at the earliest. 
More precisely, starting with the 101$^\mathrm{st}$ epoch, at the end of each epoch, we deploy Monte Carlo simulation \cite{kroese2013handbook} to determine the volume fractions and specific surface areas of phases of microstructures generated by the current model\footnote{
	Note that the specific surface area $S_V$ of some phase can be computed from a 3D image by dividing the phase's surface area \cite{schladitz2007} by the volume of the 3D image's sampling window. By assuming isotropy, the specific surface area can be computed from 2D images by considering
	$\frac{4}{\pi}L_A$, where $L_A$ denotes the specific perimeter, \ie, the perimeter of the phase in the 2D image divided by the area observed in the 2D image \cite{Chiu2013}.
}.
Then, the sum of mean absolute errors for these values is determined, where volume fractions and specific surface areas computed from training data serves as ground truth.
At the end of an epoch, we check whether this error has decreased. If no improvement is observed within 500 epochs, the training is stopped and  $\widehat{\theta}$ is set to the parameter vector that led to the smallest error with respect to volume fractions and specific surface areas.

\subsubsection{Identifying and calibrating low-parametric models}\label{sec:lpModels}
Note that the parameter vector $\theta$ of the model $\cathodeModelParametric$ has a relatively large dimension, in particular since each entry of the parameter vector $\alpha$ of an underlying kernel $\kernel_\alpha$ can be considered to be function values of the kernel evaluated on an equidistant grid, see Eqs.~(\ref{eq:kernelNonNormalized}) and (\ref{eq:isotropicKernel}). Consequently, the interpretability of individual entries of the parameter vector $\theta$ is currently not ensured. 
However, as outlined in Section~\ref{sec:introduction}, parametric models have various advantages, \eg, they allow for the systematic variation of model parameters to investigate various structural scenarios for the purpose of virtual materials testing. 
In this section we show how the  model $\cathodeModelParametric$ can be utilized to derive a low-parametric model that is suitable for generating morphologies similar to those observed in data and still can be calibrated with the methods mentioned above, in particular, with Algorithm~\ref{alg:training_procedure_gan_combined}.

\begin{algorithm}[H]
	\caption{Training approach for solving $(\ref{eq:gancombined}$) (\textbf{Inputs:} 
		$\samplingWindow$: sampling window; 
		$\widehat{C}_{ij}^\mathrm{data}$: two-point probability functions;	
		$\cathodeModelCutout$: random cutouts of image data  $S_1, \dots, S_\numSlices$ on $\samplingWindow$; $n_\mathrm{epoch}$: number  of epochs; $n_\mathrm{steps}$: steps per epoch.
		\textbf{Output:} $\widehat{\theta}$: fitted parameter vector.)}\label{alg:training_procedure_gan_combined}
	\begin{algorithmic}[1]
		\Procedure{TrainCombined}{$\cathodeModelCutout,\widehat{C}_{ij}^\mathrm{data}, W, n_\mathrm{epochs}, n_\mathrm{steps}$}
		\State $\lr \gets 0.0001$ \Comment{Learning rate model}
		\State $\lr_\mathrm{D} \gets 0.0001$ \Comment{Learning rate discriminator}
		\State $\bs \gets 32$  \Comment{Batch size}
		
		
		\State $\theta \gets \mathrm{TrainTPPF}(\widehat{C}_{ij}^\mathrm{data},\samplingWindow,1,100)$ \Comment{Initialize and pretrain generator weights for 100 steps, using Algorithm~\ref{alg:training_procedure}}

		\State initialize $\discriminatorWeights$ by simulating Gaussian noise \Comment{Weight vector}
		\For{$\mathrm{step} \text{ in } 1,\ldots,100$} \Comment{Pretrain discriminator for 100 steps}
		\State $\discriminatorWeights \gets \mathrm{DiscriminatorTrainingStep}(\cathodeModelCutout,\theta,\discriminatorWeights,\samplingWindow,\bs,\lr_\mathrm{D})$ \Comment{Discriminator pretrain based on Algorithm~\ref{alg:training_procedure_gan}}
		\EndFor

		\For{$e \text{ in } 1,\ldots,n_\mathrm{epoch}$} \Comment{Epochs}

		\For {step in $ 1,\ldots,n_\mathrm{steps}$}\Comment{Model training steps}
		\State Generate $\widetilde{\xi}_1,\dots,\widetilde{\xi}_\bs$ i.i.d. copies of 		 $\widetilde{\cathodeModel}_\theta$ on sampling window $\samplingWindow$ \Comment{Model realizations}
		\State From $\widetilde{\xi}_\ell$ compute $\widetilde{C}_{ij,\theta}^{(\ell)}$
		for each $(i,j) \in I$ and $\ell =1,\dots,\bs$ \Comment{ Eqs.~(\ref{eq:tpcpEstimator}) and~(\ref{eq:tpcp})}
		
		\State $\mathrm{loss}(\theta) \gets  \frac{1}{\bs} \sum_{\ell=1}^{\bs} (1-\discriminator_\discriminatorWeights(\widetilde{\xi}_{\ell} \vert_\samplingWindow))^2 + \frac{\gamma}{\bs} \sum_{\ell=1}^{\bs} \sum_{(i,j)\in I} \sum_{h=0}^{h_\mathrm{max}} 
		\left(
		\widetilde{C}_{ij}^\mathrm{data}(h)-\widetilde{C}_{ij,\theta}^{(\ell)}(h)
		\right)^2$	\Comment{Eq. (\ref{eq:gancombined})}				
		\State Update learning rate $\lr$ according to Adam algorithm \Comment{see \cite{Kingma2014}}
		\State $\theta \gets \theta - \lr \cdot \, \nabla_\theta \mathrm{loss}(\theta)$ 
		\EndFor
		
		\For {step in $ 1,\ldots,n_\mathrm{steps}$}\Comment{Discriminator training steps}
		\State Generate $\widetilde{\xi}_1,\dots,\widetilde{\xi}_\bs$ i.i.d. copies of 		 $\widetilde{\cathodeModel}_\theta$ on sampling window $\samplingWindow$ \Comment{Model realizations}
		\State Generate $\cathodeModelCutoutRealization_1,\dots,\cathodeModelCutoutRealization_\bs$ i.i.d. copies of 		 $\cathodeModelCutout$ \Comment{Sample data}
		\State $\mathrm{loss}_\mathrm{D}(\theta_\mathrm{D}) \gets - \frac{1}{\bs} \sum_{\ell=1}^{\bs} \discriminator_\discriminatorWeights(\cathodeModelCutoutRealization_\ell)^2+ (1-\discriminator_\discriminatorWeights(\widetilde{\xi}_{\ell} \vert_\samplingWindow))^2$	\Comment{Eq. (\ref{eq:disc_loss})}		
		\If{$\mathrm{loss}_\mathrm{D}>0.4$}	 \Comment{Avoid discriminator overfitting \cite{fuchs2024generating}}	
		\State Update learning rate $\lr_\mathrm{D}$ according to Adam algorithm \Comment{see  \cite{Kingma2014}}
		\State $\discriminatorWeights \gets \discriminatorWeights - \lr_\mathrm{D} \cdot \, \nabla_\discriminatorWeights \mathrm{loss}_\mathrm{D}(\theta_\mathrm{D})$ 
		\EndIf
		\EndFor

		\EndFor
		\State $\widehat{\theta} \gets \theta$
		\State \Return $\widehat{\theta}$
		\EndProcedure
	\end{algorithmic} 
\end{algorithm}

Therefore, let $\fittedParam$ denote the parameter vector obtained by means of Algorithm~\ref{alg:training_procedure_gan_combined}, \ie,
the model that has been calibrated by means of 2D image data is given by $\fittedCathodeModel$. Recall that this model comprises five underlying GRFs	$\widetilde{X},\widetilde{Y},\widetilde{Z},\fieldOne,\fieldTwo$, see Figure~\ref{fig:computationalScheme}, each of which has its own kernel that is parameterized by a relatively high-dimensional vector, see Eqs.~(\ref{eq:kernelNonNormalized}) and (\ref{eq:isotropicKernel}). We denote the kernel of these GRFs by $\kernel$ in this section.
From a calibrated kernel $\kernel$, by means of Eqs.~(\ref{eq:kernel2cov}) and (\ref{eq:covIso}) we can determine the corresponding covariance function $\covarianceFunction \colon [0,\infty)\to \R$ by 
$\covarianceFunction(h) = (\kernel \ast \kernel_\mathrm{mirrored})(h e_1)$ for each $h\ge 0$.
By determining a family of parametric covariance functions that can be used to adequately model the covariance functions $\covarianceFunction$ associated with the GRFs $\fieldOne,\fieldTwo,\widetilde{X},\widetilde{Y},\widetilde{Z}$, we will derive a low-parametric model for the morphology of ASSB cathodes as follows.

First, note that there are various low-parametric parametric families of covariance functions like the Cauchy family and the powered exponential family \cite{Chiu2013} as well as the  cardinal sine function \cite{lantuejoul2013}. In addition, by considering convex combinations as well as products of these covariance functions new parametric families of covariance functions can be constructed. By testing various combinations of the parametric covariance functions stated above, it turned out that the parametric covariance functions $\covarianceFunction_\alpha \colon \Z^d \to \R$ given by
\begin{equation}\label{eq:parametricCov}
	\rho_\alpha(t)= \alpha_1 \frac{\sin(\alpha_4 h)}{\alpha_4 h} e^{- \alpha_5 h^{\alpha_{11}}} + (1-\alpha_1)\begin{pmatrix}
		\alpha_2 (\alpha_3 e^{-\alpha_6 h^{\alpha_{12}}} + (1-\alpha_2)\frac{\sin(\alpha_7 h)}{ \alpha_7 h} e^{-\alpha_{8} h^{\alpha_{13}} }) + (1-\alpha_3) (1+(\alpha_{9} h)^2)^{\alpha_{10}}
	\end{pmatrix},
\end{equation}
for each $t\in \Z^d$ with $h=|t|$, defines a parametric family of covariance functions that 
is suitable for modeling the covariance functions $\covarianceFunction$ associated with the GRFs $\fieldOne,\fieldTwo,\widetilde{X},\widetilde{Y},\widetilde{Z}$---where\footnote{
	Note that according to \cite{Chiu2013} the parameter of the powered exponential kernel should have an upper boundary, \ie, the parameters $\alpha_{11},\alpha_{12},\alpha_{13}$ should be below a value of 2 such that the powered exponential kernel is a valid covariance function. We omitted this constraint on the parameters, since the model trained without constraints led to reasonably good results. 
}
$\alpha= (\alpha_1,\ldots,\alpha_{13}) \in [0,1]^3 \times (0, \infty)^{10}$. 
Then, by deploying Eq.~(\ref{eq:Cov2Kernel}) we obtain a low-parametric isotropic kernel   
\begin{equation} \label{eq:parametricKernel}
	k_\alpha= \mathrm{FFT}^{-1}\!\left( \sqrt{ \mathrm{FFT}(\rho_\alpha) } \right).
\end{equation}
This low-parametric kernel can be substituted with  
the high-parametric kernels defined in Eqs. (\ref{eq:kernelNonNormalized}) and (\ref{eq:isotropicKernel}). In particular, by substituting the previously considered high-parametric kernels with the low-parametric versions described in this section, we acquire a low-parametric model $\cathodeModel$. With an abuse of notation, we denote its low-dimensional\footnote{The number $p$ of parameters for the low-parametric model is given by $p=5 \cdot 13 + 5=70$.} parameter vector by $\theta\in \Theta \subset \R^p$, the corresponding model by $\cathodeModelParametric$ as well as the approximative model by $\cathodeModelParametricApprox$. Particularly, since the covariance function given in Eq.~(\ref{eq:parametricCov}) is differentiable with respect to its parameters, the kernel $\kernel_\alpha$ given in Eq.~(\ref{eq:parametricKernel}) and consequently the approximative, low-parametric model $\cathodeModelParametricApprox$ is differentiable.
Thus, Algorithms~\ref{alg:training_procedure} - \ref{alg:training_procedure_gan_combined} can also be used for calibrating the low-parametric model described in the present section.

As indicated by the visual comparison shown in Figure~\ref{fig:model_realization_tpp} the results achieved by the high-parametric model seem to be best, when it is calibrated by means of Algorithm~\ref{alg:training_procedure_gan_combined}. 
In order to distinguish between the high-parametric and the low-parametric models calibrated by means of Algorithm~\ref{alg:training_procedure_gan_combined}, we denote them by  $\NonParametricCathodeModel$ and $\LowParametricCathodeModel$ from here on. The goodness-of-fit of both models is investigated in Section~\ref{sec:Results2D} and discussed in  Section~\ref{sec:discussionVal}.

\subsubsection{Cylindrically isotropic ASBB cathode model}
\label{sec:Cylindircallyisotropic}

\noindent
Recall that the 3D image data depicting the microstructure of the ASSB cathodes indicates anisotropy, in particular, the  microstructure of the composite seems to be cylindrically isotropic. In other words, planar sections parallel to the $x$--$y$ plane seem to be isotropic in 2D, see Section~\ref{sec:material}.
Therefore, in order to calibrate a low-parametric but anisotropic model for ASSB cathode microstructures, we will modify an isotropic model that has been calibrated by means of planar sections parallel to the $x$--$y$ plane in the previous section.
Recalling the motivation presented in Section~\ref{sec:introduction}, low-parametric stochastic 3D models are desirable because they enable the systematic generation of a wide range of structural scenarios for the purpose of virtual materials testing. Therefore, going forward, we will only consider the low-parametric model $\LowParametricCathodeModel$ to construct a cylindrically isotropic model for the microstructure of ASBB cathodes, as this will facilitate virtual materials testing in a forthcoming study. Nevertheless, we would like to point out that the high-parametric model $\NonParametricCathodeModel$ has its merits. On the one hand, as we will demonstrate in Section~\ref{sec:Results2D}, it achieves better results than $\LowParametricCathodeModel$ due to its higher flexibility. Furthermore, it was instrumental in enabling us to select appropriate parametric covariance functions in Eq.~\eqref{eq:parametricCov}.


In order to introduce anisotropy with respect to the $z$-direction, the isotropic model $\LowParametricCathodeModel$ will be scaled into that direction. To determine a suitable scaling factor, we compute  chord length distributions  from experimentally measured image data. Formally,  the chord length distribution of a random closed set $\mathcal{C}\subset \R^3$ is defined as the length distribution of a randomly chosen line segment in $\mathcal{C} \cap \ell$, where $\ell\subset \R^3$ is a line that passes through the origin \cite{Chiu2013}. 
For details on the estimation of chord length distributions from data, \eg, from the 3D image $S$ given in Eq.~(\ref{eq:segmentation}), the reader is referred to \cite{ohser2000statistical}.

Assuming isotropy of $\mathcal{C}$, the chord length distribution is independent on the choice of $\ell$.
In particular, in the case of isotropy, the chord length distributions for lines $\ell$ parallel to the $x$-, $y$- and $z$-directions coincide. 
With this knowledge, we aim to find a global scaling factor  such that the chord length distributions estimated from the segmented image $S$ into the $z$-direction matches the distributions of chord lengths in $x$- and $y$-directions.
Therefore, let  $\Phi_{\mathrm{x},i}, \Phi_{\mathrm{y},i},\Phi_{\mathrm{z},i}: [0, \infty) \to [0,1]$ denote the distribution functions of chord lengths estimated from $S$ for the $i$-th phase in $x$-, $y$- and $z$-direction. As the experimentally measured 3D image $S$ exhibits cylindrical isotropy, the chord length distributions in  $x$- and $y$-direction are almost identical. Therefore, we determine a more robust estimation for chord length distributions parallel to the $x$--$y$ plane by considering $\Phi_{\mathrm{xy},i}= \frac{1}{2}\left( \Phi_{\mathrm{x},i} + \Phi_{\mathrm{y},i}\right)$ for each $i=1,2,3$.

Then, we compute a scaling factor $\widehat{s} \in [0, \infty)$ for the $z$-direction by solving the following minimization problem 
\begin{align}
	\widehat{s} = \underset{s \in [0, \infty)}{\text{argmin}} \int_0^\infty 
	\sum_{i=1}^{3}
	|\Phi_{\mathrm{xy},i}(t) - \Phi_{z,i}(ts)| \, \mathrm{d}t.
\end{align}
Finally, by scaling the isotropic low-parametric model $\LowParametricCathodeModel$ in $z$-direction by $\widehat{s}=0.94$, we obtain the anisotropic, low-parametric  cathode model $\anisotropicCathodeModel$, which is given by
$
\anisotropicCathodeModel(x,i)= \LowParametricCathodeModel(x_1,x_2,\lfloor x_3\widehat{s} \rceil,i)$, 
for each $i\in\{1,2,3\}$ and $x=(x_1,x_2,x_3)$ with $(x_1,x_2,\lfloor x_3/\widehat{s} \rceil)\in \samplingWindow$.

\section{Results}\label{sec:results}
In this section the spatial stochastic models, which have been calibrated to experimentally measured image data in Section~\ref{sec:modelCalibration}, will be quantitatively validated by comparing model realizations to the experimentally measured ASSB cathode microstructure. 
In particular, in Section \ref{sec:Results2D} we show results achieved by deploying Algorithm~\ref{alg:training_procedure_gan_combined} to calibrate the isotropic high- and  low-parametric 3D models denoted by $\NonParametricCathodeModel$ and $\LowParametricCathodeModel$, respectively. Recall that the experimentally measured 3D image data used for calibration purposes indicates cyllindrical isotropy with respect to the $x$--$y$ plane. Therefore, in Section~\ref{sec:Results2D} we constrain the quantitative comparison between   generated microstructures (\ie, model realizations) and experimentally measured ones to planar 2D sections.
Then, in Section~\ref{sec:Results3D} we quantitatively compare  3D microstructures generated by means of the cyllindrically isotropic model $\anisotropicCathodeModel$ and experimentally measured 3D data.


\subsection{Structural descriptors for the isotropic models}
\label{sec:Results2D}
\begin{figure}[H]
	\centering
	\begin{subfigure}[b]{0.25\textwidth}
		\centering
		\includegraphics[width=\textwidth]{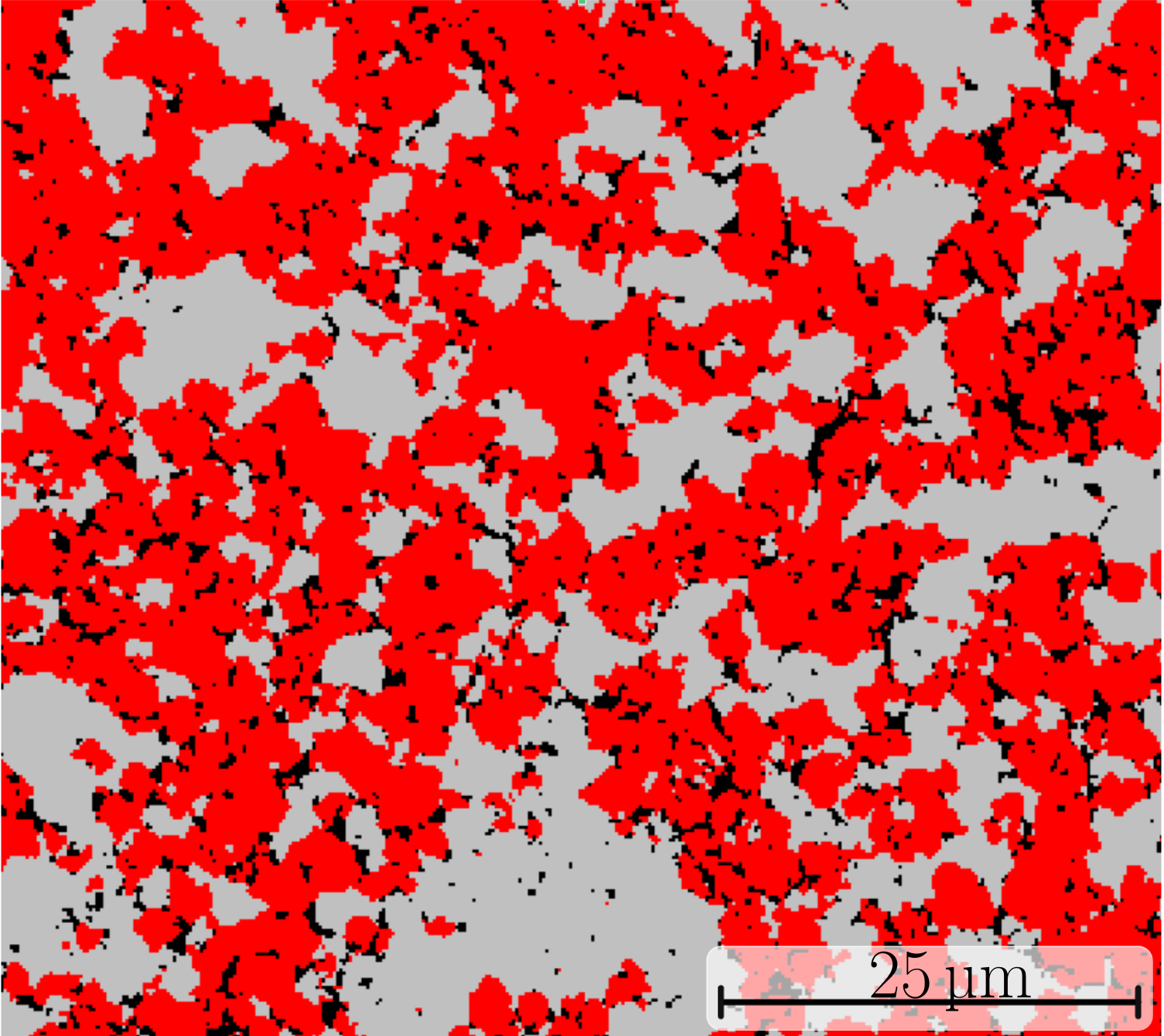}
		\caption{}		
	\end{subfigure}
	\hspace{1cm}
	\begin{subfigure}[b]{0.24\textwidth}
		\centering
		\includegraphics[width=\textwidth]{test01_2D_nonParametric_model.png}
		\caption{}		
	\end{subfigure}
	\hspace{1cm}
	\begin{subfigure}[b]{0.24\textwidth}
		\centering
		\includegraphics[width=\textwidth]{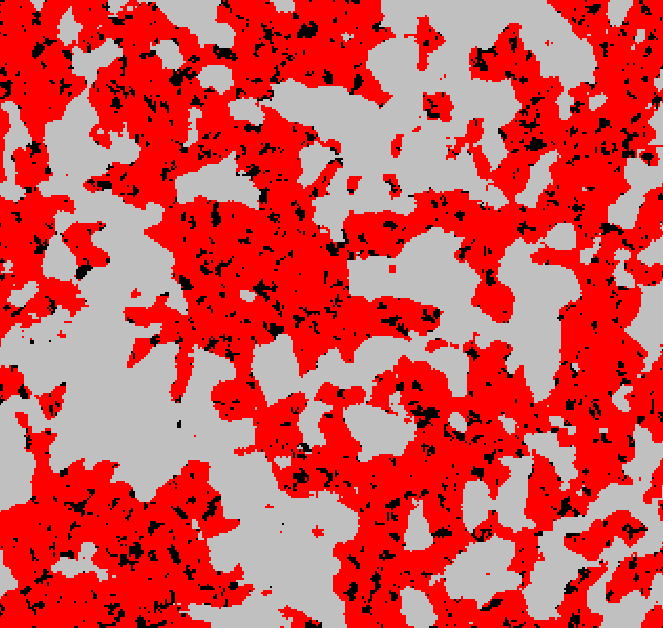}
		\caption{}		
	\end{subfigure}
	\caption{2D visualization of tomographic image data (a), and of  microstructures generated by  $\NonParametricCathodeModel$ (b) and $\LowParametricCathodeModel$ (c). The pore space, active material and the solid electrolyte are represented by black, red and gray color, respectively.  All figures use the same length scale.}
	\label{fig: 2D_img}
\end{figure}

Now we validate the isotropic models $\NonParametricCathodeModel$ and $\LowParametricCathodeModel$, that have been calibrated with Algorithm~\ref{alg:training_procedure_gan_combined}, using the 2D images as training data $S_1, \dots, S_\numSlices$. Note that, for the low-parametric model $\LowParametricCathodeModel$ it turned out that a learning rate of $0.001$ for the discriminator lead to better results.  
Planar sections of microstructures generated by the isotropic models $\NonParametricCathodeModel$ and $\LowParametricCathodeModel$ are visualized in Figures~\ref{fig: 2D_img}b and \ref{fig: 2D_img}c, whereas a planar section of the experimentally measured 3D image data is shown in Figure~\ref{fig: 2D_img}a. 

\begin{figure}[H]
	\centering
	\includegraphics[width=0.7\linewidth]{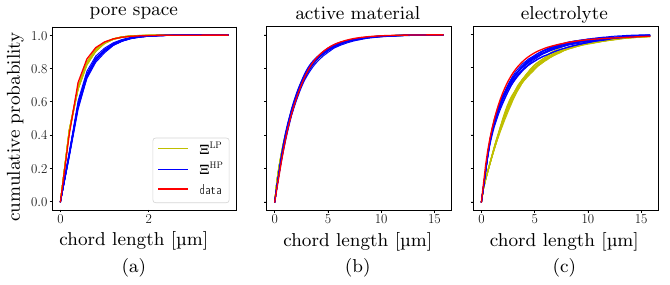}
	\caption{Chord length distribution functions computed from experimentally measured 3D image data and generated 2D microstructures for the pore space (a), the active material (b) and the electrolyte (c).
	}
	\label{fig:chord_dist_2D}
\end{figure}

In order to quantitatively evaluate the goodness-of-fit, structural descriptors are computed for ten 2D microstructures generated  by the models $\NonParametricCathodeModel$ and $\LowParametricCathodeModel$, respectively, as well as for the tomographic image data. The structural descriptors considered to compare microstructures in 2D, are phase-wise volume fractions (see Eq.~(\ref{eq:volumeFraction})), specific surface areas (see Section~\ref{sec:combined}) and mean chord lengths, where the latter are mean values associated with the chord length distributions, see Section~\ref{sec:Cylindircallyisotropic} and \cite{ohser2000statistical}. 
The resulting structural descriptors for the three phases observed in 2D sections of the experimentally measured image data and of microstructures generated by the isotropic models $\NonParametricCathodeModel$ and $\LowParametricCathodeModel$ are  listed in Table~\ref{tab:validation2D}. 
Besides the aggregated structural descriptors listed in Table~\ref{tab:validation2D}, we have determined cumulative distribution functions of the chord lengths within the phases observed in experimentally measured data as well as in generated microstructures, see Figure~\ref{fig:chord_dist_2D}.

Overall, the high-parametric model $\NonParametricCathodeModel$ outperforms the low-parametric model $\LowParametricCathodeModel$ which is no surprise since the former is more flexible, see Section~\ref{sec:discussionVal} for a detailed discussion.

\begin{table}[htb]
	\centering
	\begin{tabular}{@{}llll@{}}\toprule
		&  volume fraction & average chord length [\si{\micro\meter}] & specific surface  area[\si{\micro\meter^{-1}}]\\[2pt] \midrule
		solid electrolyte (data) & 0.42 & 2.39 & 0.66 \\
		solid electrolyte ($\NonParametricCathodeModel$) & $0.43 \pm 0.020$ & $2.68 \pm 0.169$ & $0.62 \pm 0.020$\\
		solid electrolyte ($\LowParametricCathodeModel$) & $0.42 \pm 0.015$ & $3.36 \pm 0.069$ & $0.51 \pm 0.015$\\[4pt]
		active material (data) & 0.51 & 2.06 & 0.99 \\
		active material ($\NonParametricCathodeModel$) & $0.51 \pm 0.019$ & $2.18 \pm 0.041$ & $0.96 \pm 0.025$ \\
		active material ($\LowParametricCathodeModel$) & $0.51 \pm 0.014$ & $2.13 \pm 0.067$ & $0.97 \pm 0.013$\\[4pt]
		
		pore space (data) & 0.07 & 0.40 & 0.36 \\
		pore space ($\NonParametricCathodeModel$) & $0.06 \pm 0.002$ & $0.54 \pm 0.009$ & $ 0.31 \pm 0.018$\\
		pore space ($\LowParametricCathodeModel$) & $0.06 \pm 0.001$  & $0.45 \pm 0.004$ & $0.31 \pm 0.009$\\
		\bottomrule
	\end{tabular}
	\caption{Volume fractions, specific surface areas and mean chord lengths associated with the three phases, \ie, solid electrolyte, active material and the pore space.     	Descriptors computed from planar sections of experimentally measured data are denoted by ``(data)''. Averages and standard deviations of these descriptors are determined from ten 2D microstructures generated by $\NonParametricCathodeModel$ and $\LowParametricCathodeModel$.
	}
	\label{tab:validation2D}
\end{table}


\subsection{Structural descriptors for the cylindrically isotropic model}
\label{sec:Results3D}

In order to have a low-parametric model with which we can generate 3D microstructures that mimic experimentally measured cathode microstructures, we determined the anisotropic model $\anisotropicCathodeModel$ by scaling $\LowParametricCathodeModel$  in Section~\ref{sec:Cylindircallyisotropic}.
Recall that, even though $\NonParametricCathodeModel$ seems to outperform $\LowParametricCathodeModel$, in Section~\ref{sec:Cylindircallyisotropic} we decided utilize the latter to formulate the low-parametric model $\anisotropicCathodeModel$  for ASSB cathode microstructures, as this will facilitate virtual materials testing in a forthcoming study.

\begin{figure}[ht]
	\centering
	\begin{subfigure}[b]{0.285\textwidth}
		\centering
		\includegraphics[width=\textwidth]{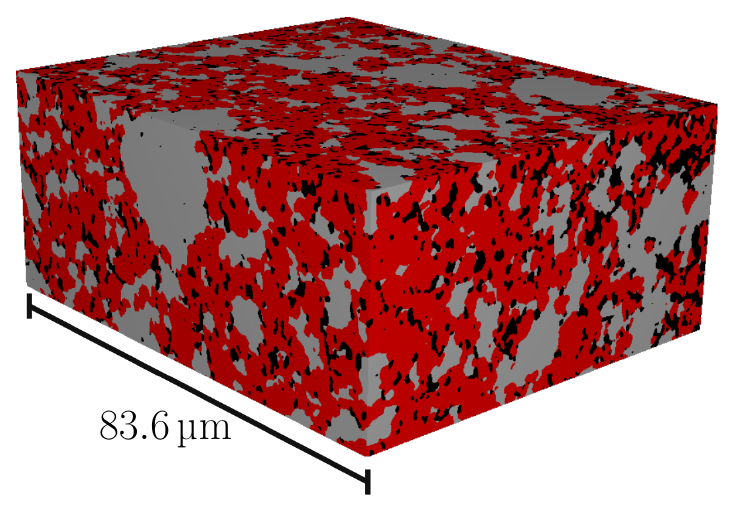}
		\caption{}		
	\end{subfigure}
	\hspace{1cm}
	\begin{subfigure}[b]{0.288\textwidth}
		\centering
		\includegraphics[width=\textwidth]{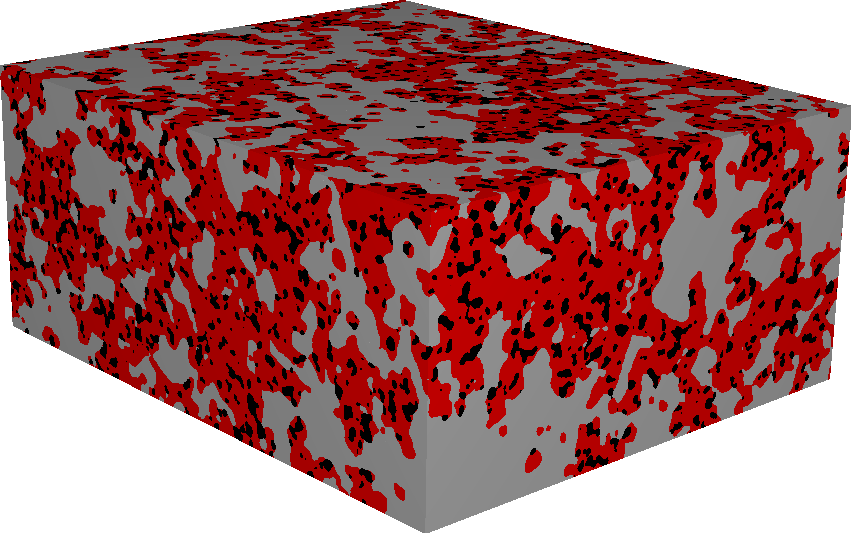}
		\caption{}		
	\end{subfigure}
	\hspace{1cm}
	\begin{subfigure}[b]{0.288\textwidth}
		\centering
		\includegraphics[width=\textwidth]{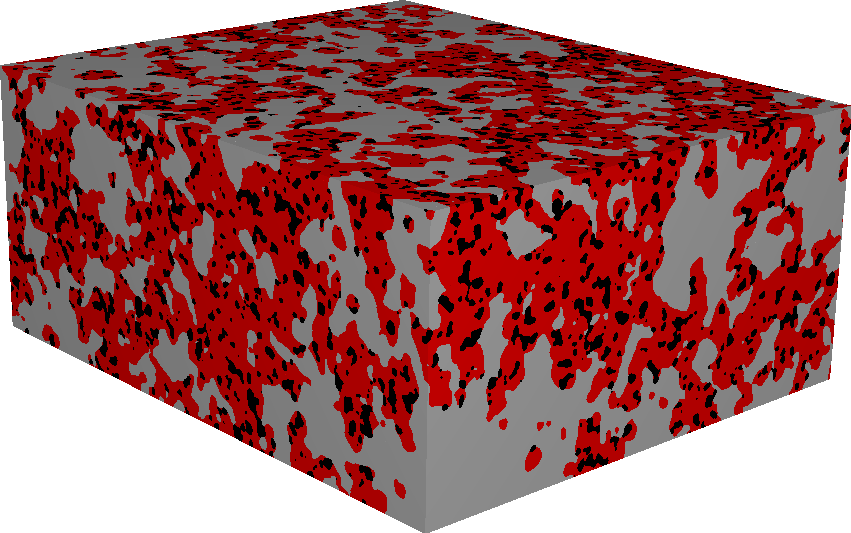}
		\caption{}		
	\end{subfigure}
	\caption{3D visualization of experimentally meausured 3D image data (a), and 3D microstructures generated by the low-parametric isotropic  $\LowParametricCathodeModel$ (b) and the anisotropic model $\anisotropicCathodeModel$ (c). The pore space, active material and the solid electrolyte are shown in black, red and gray color, respectively.  All figures use the same length scale.}
	\label{fig: 3D_img}
\end{figure}

In this section we validate this final model for the 3D microstructure of ASSB cathodes, by comparing structural descriptors computed from generated 3D microstructures to those of experimentally measured 3D image data.
To visually validate the goodness-of-fit, in Figure~\ref{fig: 3D_img}, the experimentally measured 3D image data,
as well as 3D microstructures generated by $\LowParametricCathodeModel$ and $\anisotropicCathodeModel$ are shown. 

In order to evaluate the goodness-of-fit for 
$\anisotropicCathodeModel$ quantitatively, structural descriptors are computed for 3D microstructures generated by the model as well as for the experimentally measured 3D image data. 
Similarly to the previous section, we computed the volume fractions, the mean chord lengths and the specific surface areas of the three phases observed in 3D data, where the specific surface area is computed using the method described in \cite{schladitz2007}. In addition to these descriptors, we compute further structural descriptors of the solid electrolyte and active material phase which are known to impact the macroscopic transport properties of materials---as ion and electron transport play crucial roles in the functionality of batteries.
In particular, we consider the mean geodesic tortuosity of the solid electrolyte and active material phase, which is useful to quantify the complexity of pathways from a starting plane to an end plane through the considered phase in comparison to a straight line  \cite{neumann2019}.  Notably, the mean geodesic tortuosity quantifies shorted path lengths through the transport phase. Therefore, it typically exhibits smaller values than other types of tortuosities like, for example, the medial axis tortuosity \cite{holzer2023tortuosity}. 
Moreover, as a descriptor for bottleneck effects, we consider the so-called constrictivity $\beta \in [0,1]$ of the two transport phases, see \cite{muench2008} for a formal definition. Note that values of $\beta$ close to zero indicate pronounced bottleneck effects, while $\beta = 1$ indicates no bottleneck effects. 
The considered structural descriptors have been computed for the experimentally measured 3D image data as well as for 3D microstructures generated by $\anisotropicCathodeModel$, see Table~\ref{tab:validation3D}. Note that direction-dependent descriptors (\ie, the mean geodesic tortuosity and the constrictivity) have been calculated in $z$-direction. A detailed discussion of these results is given in Section~\ref{sec:discussionVal}.
\begin{table}[htb]
	\small
	\centering
	\begin{tabular}{@{}lp{1.7cm}p{1.7cm}p{1.7cm}p{1.7cm}p{1.7cm}p{1.7cm}p{1.7cm}@{}}\toprule
		&  volume fraction & average chord length [\si{\micro\meter}] & specific surface  area[\si{\micro\meter^{-1}}] & constrictivity & mean geodesic tortuosity\\[2pt] \midrule
		
		solid electrolyte (data) & 0.42 & 2.31 & 0.64 & 0.50  & 1.11\\
		solid electrolyte ($\anisotropicCathodeModel$) & $0.45 \pm 0.006 $ & $3.80 \pm 0.066$ & $0.47 \pm 0.003$  & $0.65 \pm 0.034$ &  $1.08 \pm 0.002$\\[4pt]
		
		active material (data) & 0.51 & 2.13 & 0.87 & 0.75  & 1.07\\
		active material ($\anisotropicCathodeModel$) & $0.48 \pm 0.005$ & $1.99 \pm 0.017$ & $0.92 \pm 0.007$ & $0.85 \pm 0.050 $ & $1.07 \pm 0.001$\\[4pt]
		
		pore space (data) & 0.07 & 0.50 & 0.58 & - & -\\
		pore space ($\anisotropicCathodeModel$) & $0.08 \pm 0.001$ & $0.49 \pm 0.002 $ & $0.57 \pm 0.009$  & - & -\\
		\bottomrule
	\end{tabular}
	\caption{Structural descriptors computed for experimentally measured 3D image data. Averages and standard deviations of the descriptors considered for the model, $\anisotropicCathodeModel$, were computed based on ten 3D microstructures.}
	\label{tab:validation3D}
\end{table}

\begin{figure}[htb]
	\centering
	\includegraphics[width=0.8\linewidth]{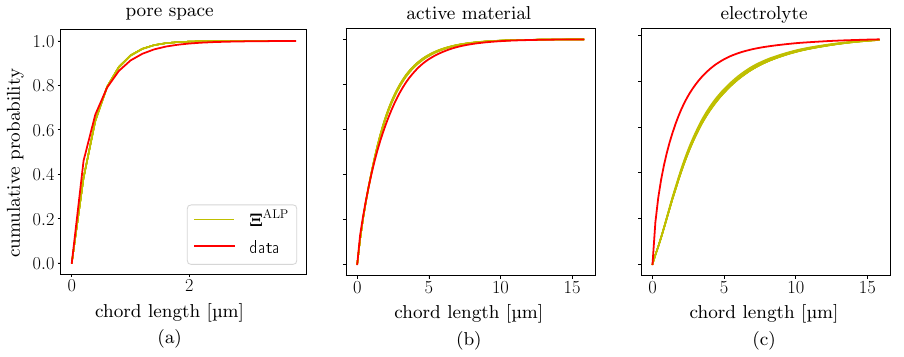}
	\caption{Chord length distribution functions computed from experimentally measured 3D image data 
		and  3D microstructures drawn from $\anisotropicCathodeModel$ associated with  pore space (a),   active material (b) and   solid electrolyte~(c).}
	\label{fig:chord_dist_3D}
\end{figure}

Besides these aggregated structural descriptors, we also consider the cumulative distribution functions of chord lengths in $z$-direction, see Figure~\ref{fig:chord_dist_3D}.


\section{Discussion}\label{sec:discussion}
In this section we discuss the methods and results presented in this paper. In particular, in Section~\ref{sec:discMethod} we start with a  discussion on the computational methods introduced in the present paper. Then, in Section~\ref{sec:discussionVal} the results presented in Section~\ref{sec:results} are discussed.

\subsection{Generative adversarial calibration}\label{sec:discMethod}
The excursion set models proposed in this paper build on the work presented in \cite{neumann2019}, where excursion sets of correlated Gaussian random fields have been used to model the microstructure of a three-phase material. Here, we extend this approach by using correlated $\chi^2$-random fields, which are constructed from correlated Gaussian random fields with a flexible correlation structure, as defined in Eq.~(\ref{eq:correlatedFields}). This added flexibility enhances our model's adaptability to various microstructural patterns but comes at the cost of increased model complexity.

In \cite{neumann2019}, model calibration could be performed, using analytical formulas that related volume fractions and two-point coverage probability functions observed in image data with parameters of the model. To our knowledge these analytical formulas do not hold for the proposed models of the present paper. To overcome this challenge, we have developed a computational method that facilitates the calibration of our flexible models to volume fractions and two-point coverage probability functions, see Algorithm~\ref{alg:training_procedure}.
The results shown in Figure~\ref{fig:model_realization_tpp}a, however, indicate that a calibration by means of Algorithm~\ref{alg:training_procedure} results in a model that generates microstructures with increased surface roughness. 
An explanation for this could be that there may exist multiple configurations for the model parameters that lead to adequately fitting descriptors (two-point coverage probability functions and volume fractions), making it challenging to uniquely identify well-suited parameters. Therefore, motivated by the framework of GANs, in Algorithm~\ref{alg:training_procedure_gan} we propose an alternative calibration approach that utilizes a neural network that serves as a discriminator. Roughly speaking the discriminator is supposed to learn descriptors that enable it to distinguish between generated and experimentally measured microstructures---thus, during calibration the model tries to minimize the discrepancy of the descriptors learned by the discriminator. Note that there are many reports in the literature referring to difficulties when training GANs \cite{saad2024survey}, \eg, it can occur that the discriminator learns descriptors that do not provide meaningful feedback to the model to be trained \cite{saad2024survey}. This could explain the large discrepancies between experimentally measured image data and  microstructures generated by the model that has been trained with Algorithm~\ref{alg:training_procedure_gan}, compare Figures~\ref{fig:anistropy}b and~\ref{fig:model_realization_tpp}b.

To overcome these challenges, we combined both calibration approaches. In Algorithm~\ref{alg:training_procedure_gan_combined}, the model is calibrated to minimize descriptors learned by a discriminator while also ensuring that the two-point coverage probability functions of the generated microstructures align with those observed in experimentally measured image data. In direct comparison with the other algorithms, this third approach, as presented in Algorithm~\ref{alg:training_procedure_gan_combined}, yields the best visual results, see Figure~\ref{fig:model_realization_tpp}c. 
An explanation for this improvement could be that training with  two-point coverage probability functions already leads to promising results, see Figure~\ref{fig:model_realization_tpp}a. Consequently, the discriminator is trained to distinguish between experimentally measured data and already quite realistically looking model realizations. Thus, we believe that the discriminator is able  
to learn additional, more meaningful descriptors that can guide model calibration.


\subsection{Model validation}\label{sec:discussionVal}
In this section we discuss the results achieved by the isotropic high- and low-parametric models $\NonParametricCathodeModel$ and $\LowParametricCathodeModel$. In particular, Table~\ref{tab:validation2D} indicates that the volume fractions observed in data nearly coincides with volume fractions of microstructures generated by $\NonParametricCathodeModel$ and $\LowParametricCathodeModel$. In general, the mean chord lengths and specific surface areas are well reproduced by the models for the active material and the pore space, even if some values are slightly over- or underestimated.  
Overall the structural validation of $\NonParametricCathodeModel$ and $\LowParametricCathodeModel$ by means of Table~\ref{tab:validation2D} and Figure~\ref{fig:chord_dist_2D} indicates that generated 2D microstructures of these isotropic models fit reasonably well to planar sections of the experimentally measured 3D image data. 
In comparison to $\NonParametricCathodeModel$, the low-parametric model $\LowParametricCathodeModel$ shows larger deviations with respect to the specific surface area and mean chord length of the solid electrolyte---where a decrease in goodness-of-fit is to be expected since the high-parametric model $\NonParametricCathodeModel$ is more flexible, \ie, more general than the low-parametric version $\LowParametricCathodeModel$. Still we have chosen to pursue results of the low-parametric model $\LowParametricCathodeModel$ as it will facilitate virtual materials testing in future research, as outlined in Section~\ref{sec:introduction}.   
However, it is important to emphasize the merits of the high-parametric model, $\NonParametricCathodeModel$. Its increased flexibility not only yields better results but also plays a crucial role in guiding the selection of suitable parametric covariance functions in Eq.~\eqref{eq:parametricCov} for constructing $\LowParametricCathodeModel$.


Now we discuss the results obtained by scaling  $\LowParametricCathodeModel$ into $z$-direction, to mimic the anisotropy observed in experimentally measured 3D image data---resulting in the model $\anisotropicCathodeModel$. 
The 3D visualization in Figure~\ref{fig: 3D_img} of experimentally measured image data and a 3D microstructure generated by $\anisotropicCathodeModel$ indicates a good fit.
The goodness-of-fit is mostly confirmed by the results shown in Table~\ref{tab:validation3D} as well as in Figures~\ref{fig:chord_dist_3D}, 
with the exception of some larger deviations for descriptors associated with the solid electrolyte phase. These errors shown in Table~\ref{tab:validation3D} presumably stem from the underlying, unscaled model $\LowParametricCathodeModel$ listed in Table~\ref{tab:validation2D}, where we observed  similar discrepancies due to additional model assumptions on covariance functions in Eq.~(\ref{eq:parametricCov}).
In comparison, to related models that have been fitted to image data of three-phase microstructures in the literature, we observe similar relative errors, see, \eg, \cite{neumann2019pluri}. 
Given the complexity of the microstructure of ASSB models, we  believe that the results achieved by the low-parametric model $\anisotropicCathodeModel$ are satisfactory, which motivates the deployment of this model for the purpose of virtual materials testing in future research.

\section{Conclusions}\label{sec:conclusions}
We developed a computational framework for computing digital twins for the complex 3D morphologies of multiphase materials, using excursion sets of random fields.  These digital twins can be utilized to generate virtual but realistic 3D microstructures.
Common parametric models of stochastic geometry (\eg, excursion sets of Gaussian random fields) allow for the generation of realistic, yet unobserved, structures through systematic parameter variation. However, as model complexity increases---which is necessary to capture more intricate microstructures---the required number of model parameters increases substantially. On the other hand classical (non-parametric) GANs offer a data-driven approach for modeling complex 3D morphologies. However, after calibration the systematic variation of their model parameters for generating diverse, not yet measured structural scenarios can be difficult. Yet, the latter is of 
particular interest for virtual materials testing schemes---by  systematic variations of model parameters a wide spectrum of structural scenarios can be investigated, such that the corresponding digital twins can be exploited as geometry input for numerical simulations of macroscopic effective properties. 
The method proposed in the present paper allows for the calibration of parametric excursion set models with increased complexity by combining methods from stochastic geometry with generative AI, \ie, we utilize so-called two-point coverage probability functions as well as discriminators that are commonly used for training GANs. Under the assumption of isotropy, the model for the 3D morphology  of multiphase microstructures can even be calibrated using 2D image data.

The  proposed calibration method has been deployed to fit a parametric digital twin for the 3D microstructure of ASSB cathode materials. In future research, the model will be deployed for virtual materials testing. Therefore, we will investigate how perturbations to fitted parameters influence the structure of generated 3D morphologies.
Therefore, we will calibrate the parametric digital twin using additional image data of ASSB cathode materials with differently sized solid electrolyte particles, as investigated in \cite{Minnmann_2024}. In doing so, we aim to identify additional parameters within the model's parameter space. This will enable us to develop strategies for perturbing model parameters to explore new structural scenarios.
More precisely, we will generate a broad spectrum of differently structured 3D morphologies on which effective macroscopic properties of the ASSB cathode materials will be simulated by numerical computations---in this manner, we will derive structure-property relationships  that can guide the design of new ASSB cathode materials. The effective macroscopic properties will be simulated with the software tool GeoDict for which validated simulation models for ASSB electrodes are in development.




\section*{Data availability}
The data utilized for modeling is sourced from the study \cite{Minnmann_2024} and can be accessed via \url{http://dx.doi.org/10.22029/jlupub-18458       }.

\section*{Code availability}
All formulations and algorithms necessary to reproduce the results of this study are described in the Results and
Methods sections.

\if\submissionmain1

\else
\section*{Acknowledgements}
This research is funded by the German Federal Ministry of Education and Research (BMBF) under  grant number 03XP0562B.
The responsibility for the content of this publication lies with the authors.


\fi

\section*{Competing interests}
The authors declare no competing financial or non-financial interests.

\end{document}